%% file: main_arxiv.tex
\documentclass{article}

\usepackage{arxiv}

\usepackage[utf8]{inputenc} 
\usepackage[T1]{fontenc}    
\usepackage{booktabs}       
\usepackage{amsfonts}       
\usepackage{nicefrac}       
\usepackage{microtype}      
\usepackage{lipsum}		
\usepackage{graphicx}
\usepackage{natbib}

\usepackage{doi}
\usepackage{float}

\usepackage{amsmath}
\usepackage{amssymb}
\usepackage{mathtools}
\usepackage{amsthm}
\usepackage{pifont}
\usepackage{subcaption}

\usepackage[textsize=tiny]{todonotes}

\newcommand{\bfu}[1]{\textbf{\underline{#1}}}

\renewcommand{\cite}[1]{\citep{#1}}

\newcommand{\cmark}{\textcolor{green!80!black}{\ding{51}}}
\newcommand{\xmark}{\textcolor{red}{\ding{55}}}

\usepackage{hyperref}

\usepackage{xcolor}
\hypersetup{colorlinks=true, citecolor=blue}

\makeatletter
\renewcommand\hyper@natlinkbreak[2]{#1}
\makeatother

\date{}

\title{ArchesWeather \& ArchesWeatherGen: a deterministic and generative model for efficient ML weather forecasting}

\author{ {Guillaume Couairon}\thanks{Corresponding author} \\
	INRIA, France \\
	\texttt{guillaume.couairon@gmail.com} \\
	\And
	{Renu Singh} \\
	INRIA, France \\
	\texttt{renu.singh@inria.fr} \\
        \And
        {\hspace{1mm}Anastase Charantonis} \\
	INRIA, France \\
	\texttt{anastase.charantonis@inria.fr} \\
        \AND
	{\hspace{1mm} Christian Lessig} \\
	ECMWF, Germany\\
	\texttt{christian.lessig@ecmwf.int} \\
        \And
        {\hspace{1mm}Claire Monteleoni} \\
	INRIA, France \\
	\texttt{cmontel@inria.fr} \\
}

\begin{document}

\maketitle
\input{sections/core}

\bibliography{bib, bib2}

\bibliographystyle{icml2024}

\newpage
\input{sections/appendix}

\end{document}

%% file: sections/core.tex
\newcommand{\x}{\mathbf{x}}
\newcommand{\z}{\mathbf{z}}
\renewcommand{\r}{\mathbf{r}}

\begin{abstract}

Weather forecasting plays a vital role in today's society, from agriculture and logistics to predicting the output of renewable energies, and preparing for extreme weather events. Deep learning weather forecasting models trained with the next state prediction objective on ERA5 have shown great success compared to numerical global circulation models. However, for a wide range of applications, being able to provide representative samples from the distribution of possible future weather states is critical. In this paper, we propose a methodology to leverage deterministic weather models in the design of probabilistic weather models, leading to improved performance and reduced computing costs. We first introduce \textbf{ArchesWeather}, a transformer-based deterministic model that improves upon Pangu-Weather by removing overrestrictive inductive priors. We then design a probabilistic weather model called \textbf{ArchesWeatherGen} based on flow matching, a modern variant of diffusion models, that is trained to project ArchesWeather's predictions to the correct distribution of ERA5 weather states. ArchesWeatherGen is a true stochastic emulator of ERA5 and surpasses IFS ENS and NeuralGCM on all WeatherBench headline variables (except for NeuralGCM's geopotential).
Our work also aims to democratize the use of deterministic and generative machine learning models in weather forecasting research, with academic computing resources. All models are trained at 1.5º resolution, with a training budget of $\sim$9 V100 days for ArchesWeather and $\sim$45 V100 days for our best ArchesWeatherGen model. For inference, ArchesWeatherGen generates 15-day weather trajectories (with 24h time steps) at a rate of 1 minute per ensemble member on a A100 GPU card. To make our work fully reproducible, our code and models will be open source, including the complete pipeline for data preparation, training, and evaluation, at \url{https://github.com/INRIA/geoarches}.


\end{abstract}

\section{Introduction}

The field of weather forecasting is undergoing a revolution. AI models trained on the ERA5 reanalysis dataset \cite{hersbach2020era5} can now outperform IFS-HRES (the reference numerical weather prediction model developed by ECMWF, the European Center for Medium-Range Weather Forecasting) in a wide range of scores \cite{bi2022pangu, pathak2022fourcastnet, lam2022graphcast, chen2023fuxi, nguyen2023scaling, guo2024fourcastnext, kochkov2023neural}. At the same time, computing costs for making a forecast are orders of magnitude lower.
The standard procedure for training these machine learning (ML) based weather models has been to fit a neural network to predict the next atmospheric state given the current one, with a lead time of 6 to 24 hours.  The neural network is then used autoregressively to make forecasts at longer lead times, through time-stepping.

\paragraph{Architecture.}
The neural network architectures of machine learning weather models are typically adopted from the computer vision community, usually by adding priors related to the specificity of processing physical fields from a 3D spherical atmosphere (local 3D attention for Pangu-Weather \cite{bi2022pangu}; Fourier Spherical Operators for FourCastNet \cite{pathak2022fourcastnet}; Graph Neural Networks on a spherical mesh for GraphCast \cite{lam2022graphcast}). Adding these physical priors usually serves two goals: (i) AI models that have more priors are more interpretable since they more closely relate to their numerical counterparts, which increases trust in these models; (ii) networks with more physical priors generalize better and can reach the same accuracy with fewer parameters and memory footprint.

However, recent work has started questioning this second assumption, showing that architectures with less physical priors can also generalize well with a lower training cost \cite{nguyen2023scaling, chen2023fuxi, lessig2023atmorep}, which could mean that models with more physical priors and fewer parameters are harder to train. These works have adapted vision transformers \cite{dosovitskiy2020image}, considering ERA5 as latitude/longitude images, and concatenating upper-air weather variables in the channel dimension. This concatenation requires a lot more parameters than 3D processing, so these works still rely on very large neural networks (300M parameters for Stormer, 1.5B for FuXi).

In this paper, we identify a limitation of 3D local attention, used in the Pangu-Weather architecture. Inside the network, only the features for neighboring pressure levels interact, mimicking the physical principle that air masses only interact locally at short timescales. We find that despite its connection to physics, this prior is computationally suboptimal, and we design a global \textbf{C}ross-\textbf{L}evel \textbf{A}ttention-based interaction layer (CLA in short) that improves forecasting skill.

\paragraph{Limitations of deterministic models.} State-of-the-art machine learning models reach a lower root mean squared error (RMSE) compared to models based on physical simulation, at the cost of producing smoother outputs that lack physical consistency at short (<300km) spatial scales \cite{bonavita2024some}. Furthermore, the smoothing compounds over time, as the smoothed outputs are reused as inputs for predicting weather trajectories. Fine-tuning the models for auto-regressive rollouts helps to reduce RMSE but increases smoothing even more, resulting in forecasts in between a realistic physical simulation and a smooth ensemble mean.
This smoothing phenomenon has several drawbacks: (i) extreme events are not well represented by the forecast model; (ii) the ensemble of ML predictions does not cover the range of plausible physical scenarios, which is critical to deal with the chaotic nature of weather in the face of initial conditions uncertainty; (iii) smooth outputs that lack physical realism are less useful in some downstream applications that require physical interpretability.

Solving these issues requires sampling from the true probability distribution of possible weather trajectories given our best knowledge of the initial condition. The distribution of possible weather trajectories can be approximated by combining classical methods with deterministic machine learning models, for example, generating an ensemble from perturbed initial conditions \cite{bi2022pangu}. While the coverage of the resulting trajectories improves, the dynamics of the model given the input is unchanged; more importantly, this method does not address smoothing as the model was trained to minimize MSE and not to generate true weather states from the data distribution.
We propose solving the smoothing issue by using generative modeling.

\paragraph{Generative models for weather forecasting.} Generative modeling refers to a class of machine learning techniques that, given a sample dataset, learn to model the underlying probability distribution of the data. This allows one to generate new synthetic data samples without any assumption on the probability distribution. Generative models can learn conditional distributions, i.e. a mapping from labels to distributions. A popular conditional generation task is text-to-image: given an input text, the goal is to generate an image that is correctly described by that text \cite{ramesh2021zero, esser2021taming, saharia2022photorealistic, rombach2022high}. In the context of weather forecasting, ERA5 is a collection of weather states that can serve as a dataset of samples for generative modeling. Our goal is to generate weather trajectories from a given initial condition, so the natural framework is to use a weather state as conditional input, and to model the set of possible trajectories given this input. Since ERA5 represents a single historical trajectory, given an initial condition, there is only a single sample of the conditional distribution, formed by the states following the initial condition. However, the historical ERA5 dataset from 1979 to present provides enough samples to learn a conditional model with only one sample per initial condition, meaning that the model learns to approximate the true underlying probabilistic mapping and can generalize to other initial conditions.

Direct modeling of the distribution of weather trajectories allows us to better cover the range of future weather events compared to using deterministic ML models. Furthermore, well trained generative weather models do not suffer from the smoothing issue of deterministic models, since smoothed outputs are not correct samples from the (conditional) data distribution.
In theory, sampling a distribution over trajectories requires a neural network to predict a full trajectory at once, which is impractical since deterministic ML weather models have been designed to predict individual weather states. However, we can use a Markovian approximation of the true dynamics, where the distribution of the next weather state only depends on the current state. Learning the generative model then amounts to learning the \textit{transition distribution} of the next state given the input state. Trajectories for longer lead times are then sampled auto-regressively by sampling each transition in the trajectory. We choose this approach for ArchesWeatherGen which is also followed in GenCast \cite{price2023gencast} (see Section \ref{related_work} for more details).

While generative models allow one to model the fundamental uncertainty in weather dynamics, atmospheric processes are still nearly deterministic at short timescales (e.g., 6 to 24 hours), which partly explains the success of deterministic ML models trained for this time scale. The resulting transition distribution for these lead times is narrowly centered around its expectation, reflecting the small uncertainty. Therefore, a large part of the generative training consists of learning this expectation, which is exactly the training objective of deterministic models. 
This naturally calls for a decomposition of training into two stages: first, learning the deterministic component with the standard RMSE objective, which approximates the distribution mean; second, subtract this deterministic component from the training data and train a generative model on the resulting data, which we call \textit{residual data} or \textit{residuals}. This approach was first proposed in the context of superresolution \cite{shang2024resdiff} and successfully applied to generative downscaling \cite{mardani2309residual}. In this paper, we show that this strategy is computationally much more efficient than generative modeling from scratch on the transition distribution. We also investigate the difficulties that arise from this two-stage approach, which are mainly due to the overfitting of the deterministic model.

\begin{figure}[t]

    \vspace{-2em}
    \begin{subfigure}{.6\linewidth}
    \centering
    
    \includegraphics[width=\linewidth]{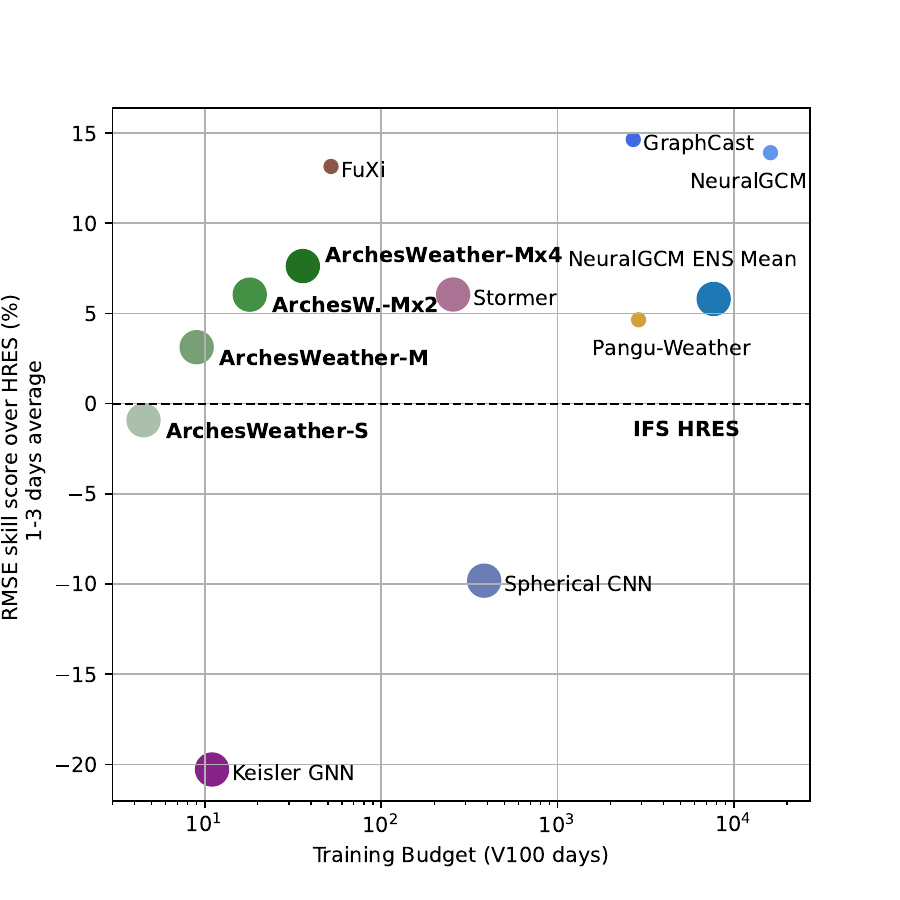}
    \end{subfigure}%
    \begin{subfigure}{.4\linewidth}
    \centering
    
    \hspace*{-0.05\linewidth}\includegraphics[width=1.1\linewidth]{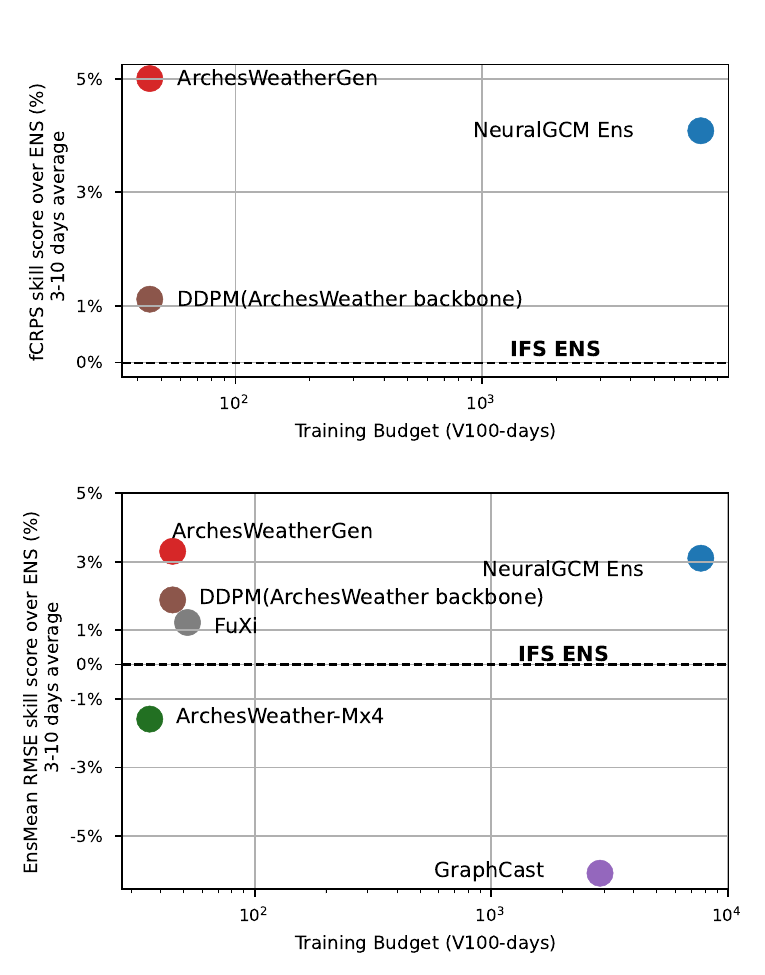} 
    \end{subfigure}
    \caption{Summary evaluation metrics (higher is better) on key upper air variables (Z500, Q700, T850, U850 and V850) as a function of training computational budget, comparing ArchesWeather and ArchesWeatherGen to other deterministic and ensemble-based models.
    \textbf{Left}: RMSE skill score over IFS HRES averaged for lead times of 1 to 3 days, comparing ArchesWeather to other state-of-the-art deterministic ML models in WeatherBench2. Circle size indicate training resolution: small circles for 0.25º/0.7º, big circles for 1º/1.4º/1.5º. Compared to the other models trained at a similar resolution, ArchesWeather reaches competitive or better forecasting performance with a much smaller training budget.
    \textbf{Right:} Ensemble metrics skill scores over IFS ENS, averaged for lead times of 3 to 10 days. \textbf{Upper Right:} ArchesWeatherGen reaches better fair CRPS scores (fCRPS, see section \ref{sec:evaluation}) than NeuralGCM at a much lower computational budget, and our flow-matching based design greatly improves upon the original DDPM diffusion models. \textbf{Lower Right:} ArchesWeatherGen also better approximates the Ensemble Mean than competing methods, including the deterministic model FuXi that was explicitly trained to match the ensemble mean. GraphCast is shown as reference for a non ensemble-based deterministic model, but was not trained to optimize this metric at 3-10 days.}
    \label{fig:teaser}
    \vspace{-0.5em}
\end{figure}

We use diffusion-based models \cite{ho2020denoising, karras2022elucidating} as our generative modeling framework, inspired by their huge success in text-to-image generation \cite{rombach2022high, saharia2022photorealistic}. These models are trained as denoisers: samples from the data distribution are mixed with Gaussian noise at various signal-to-noise ratios, and a neural network is trained to predict the original samples. New samples can be generated by sampling random Gaussian noise and iteratively denoising it with the neural network. The advantage of diffusion-based models compared to generative adversarial networks (GANs) is believed to be due to the denoising loss function, which provides stable learning and is directly correlated with sample quality \cite{esser2024scaling}. In this work, we choose a variant of diffusion models called flow matching models \cite{lipman2022flow}, which we show to perform experimentally better than the original version of diffusion models, DDPMs \cite{ho2020denoising, song2020denoising}.

\paragraph{Summary of results and contributions.} 

\begin{itemize}
    \item \textit{We show that the 3D local attention in Pangu-Weather is computationally suboptimal, and we design a nonlocal cross-level attention layer that boosts performance.} We train our deterministic model ArchesWeather at 1.5º resolution and 24h lead time. ArchesWeather achieves competitive RMSE scores with a computational budget orders of magnitude smaller than competing architectures (see Figure \ref{fig:teaser}). An ensemble average of 4 models is competitive with the 1.4º NeuralGCM ensemble with 50 members \cite{kochkov2023neural} for lead times up to 3 days.
    \item \textit{We show that we can leverage deterministic weather models to train generative weather models with improved performance and reduced computational costs.} We use our ArchesWeather models to remove the predictable component from ERA5 and train flow matching models on the residual data, with the same neural network architectures. We show that flow matching models \cite{lipman2022flow} perform better than the original version of diffusion models, DDPMs \cite{ho2020denoising, song2020denoising}. The resulting model ArchesWeatherGen outperforms NeuralGCM on temperature, humidity, and wind components, but not geopotential, as measured by probabilistic forecasting metrics: EnsembleMean RMSE, CRPS, and Brier score.
    \item  \textit{Our work allows for ML-based weather forecasting research with academic resources.} ArchesWeather requires only 1 TB of data to download (versus 36 times more at 0.25º), a training budget of $~5$ A100-days and has an inference cost of $~0.25s$ for a 24h forecast on the same hardware. Our best generative model ArchesWeatherGen requires a computational budget of $~23$ V100 days (mainly allocated to training four deterministic ArchesWeather models), and inference is longer due to the iterative denoising mechanism of diffusion-based models, which requires $3.5s$ per 24h forecast on an A100 card. Our code and models will be released open-source, including the full pipelines for data preparation, training, and evaluation.
\end{itemize}

\section{Related Work} \label{related_work}

In this work, we mainly compare ArchesWeatherGen to NeuralGCM \cite{kochkov2023neural}, since its output is available in WeatherBench for evaluation. NeuralGCM is a hybrid physics-machine learning method, with a dynamical core for numerical simulation, and a neural network trained to adjust the simulation to minimize the MSE forecast error.

The most similar work to ours is GenCast, \cite{price2023gencast}, which is also based on diffusion models for ensemble weather forecasting. In addition to requiring less resources to train (since we operate at 1.5º instead of 0.25º), the most notable differences are the architecture (original Transformer \cite{vaswani2017attention} for GenCast, Swin vision transformer \cite{liu2022swin} for ours), the diffusion paradigm (Elucidated Diffusion Model \cite{karras2022elucidating} versus Flow matching \cite{lipman2022flow}), design choice to increase diversity (EDA perturbations for GenCast, noise scaling, and OOD fine-tuning for us). In particular, we use flow matching, which is a more recent variant of diffusion, easier to work with, and has shown excellent results in image generation \cite{esser2024scaling}.
Our model also requires much less computational resources for training (about 20 times less including our deterministic model training). We do not compare our model with GenCast since, at the time of writing, the inference output of GenCast is not available for evaluation.
Other works have incorporated diffusion models for global weather forecasting, such as SwinRDM \cite{chen2023swinrdm} and Fuxi-extreme \cite{zhong2024fuxi}. We did not evaluate these models because they are not publicly available. Diffusion models can also be used to generate more ensemble members from a base set of physically-simulated members, e.g. \cite{li2023seeds, brochet2023using}.

Diffusion models have also been successful in atmospheric science tasks that require generating stochastic, high-frequency fields. One first example is downscaling \cite{mardani2024residual, srivastava2023probabilistic, watt2024generative, tomasi2024can, han2024generative, lopez2024dynamical}.
Another natural application of diffusion models in atmospheric sciences in precipitation nowcasting, due to the high stochasticity of the field \cite{yu2024diffcast,zhao2024advancing, gong2024cascast, addison2024machine, guilloteau2024generative}. Other fields that have benefitted from diffusion-based approaches are tropical cyclones forecasting \cite{nath2023forecasting, huang2024tcp}, sea ice modeling \cite{finn2024towards}.
Finally, diffusion-based models are not only competitive in probabilistic modeling, they are also well fitted for data assimilation and have been adapted for this purpose \cite{huang2024diffda, rozet2023score, manshausen2024generative}.

\section{Methods}

We first present our problem setup and proposed neural network architecture; we then cover our methodology for training and evaluating both deterministic and generative weather models.

Our objective is to predict the evolution of weather variables in the ERA5 dataset \cite{hersbach2020era5} regridded to 1.5º resolution, which is the standard used for evaluation at the World Meteorological Organization (WMO). Following the standard in Weatherbench 2 \cite{rasp2023weatherbench}, we use 1979 to 2018 as training period and test our models at 00/12UTC for each day of 2020. For our generative model, we sometimes use 2019 as a fine-tuning period. We consider standard hyper-parameters for training and hence do not use a validation set.

A weather state at time $t$ is noted $\x_t \in \mathbb{R}^{n}$ and consists of 6 upper air variables (temperature, geopotential, specific humidity, wind components U, V and W) sampled on a latitude-longitude grid at 13 pressure levels, and 4 surface variables (2m temperature, mean sea level pressure, 10m wind U and V). $\mathcal{D} = (\x_t)_{t\in \mathcal{T}}$ is the historical trajectory of weather variables, sampled every 6h. Given an input state $\x_t$, we aim to predict the future trajectory $(\x_{t+k \delta})_{1\leq k \leq K}$, where $\delta$ is the lead time, which is set to $\delta=24h$ for the remainder of the paper. Both our deterministic and generative weather models require one to map an input state $\x_t$ to an estimation of $\x_{t+\delta}$. In the next section, we present our design of neural architecture for this task.

\subsection{Architecture}

Our neural network architecture is a 3D Swin U-Net transformer  \cite{liu2021swin, liu2022swin} with the Earth-specific positional bias, largely inspired by the Pangu-Weather architecture \cite{bi2022pangu} which first proposed this design choice. The surface and upper-air variables are first embedded into a single tensor of size $(d, Z, H, W)$ where $d$ is the embedding dimension, $Z$ the vertical dimension, $H$ and $W$ the latitude and longitude dimensions. This tensor is then processed by the U-Net transformer and is projected back to surface and upper-air variables at the end.

\paragraph{Local 3D attention in Pangu-Weather.} The attention scheme in a Swin layer \cite{liu2021swin} consists of splitting the input tensor into non-overlapping windows, where a self-attention layer processes data in each window independently. The input tensor is then shifted by half a window to compute the next self-attention layer, allowing interaction between the different attention windows. In Pangu-Weather, the input tensors are split into 3-dimensional windows of size $(Z_{window}=2, H_{window}=6, W_{window}=12)$: hence, along the vertical $Z$ dimension, only the features for neighboring pressure levels interact, mimicking the physical principle that air masses only interact locally at short timescales. This inductive prior is meant to have the neural network roughly reproduce physical interaction phenomena and reduce the number of parameters needed.

From a computational perspective, this prior is a limitation since computations for similar phenomena happening at different atmospheric layers are performed independently in parallel. Global vertical interaction would allow sharing of such computations and allocating resources more efficiently. Computations for complex variables can also be spread across levels faster, to reach lower error. Finally, from a physical perspective, having vertical interaction can allow us to detect the vertical profile of the atmosphere and to adjust computations accordingly.

Before presenting our proposed solution, we mention two other potential methods and their caveats. First, one could increase the attention window size, e.g. to $(4, 6, 12)$ instead of $(2, 6, 12)$, to accelerate exchange of information along the vertical dimension, but this decreases inference speed due to the quadratic cost of attention in the sequence length. Second, some works use a more standard 2D transformer \cite{nguyen2023scaling, chen2023fuxi} and stack variables across pressure levels in a single vector at each spatial position. This comes at the cost of an increased parameter count: With $Z$ pressure levels (after embedding), the linear and attention layers need $O(d^2Z^2)$ parameters, with $d$ being the embedding dimension for a single pressure level. As a result, Stormer uses a ViT-L with 300M parameters and FuXi uses a SwinV2 architecture with 1.5B parameters.

\begin{figure}[ht]
    \centering
    \includegraphics[width=0.95\linewidth]{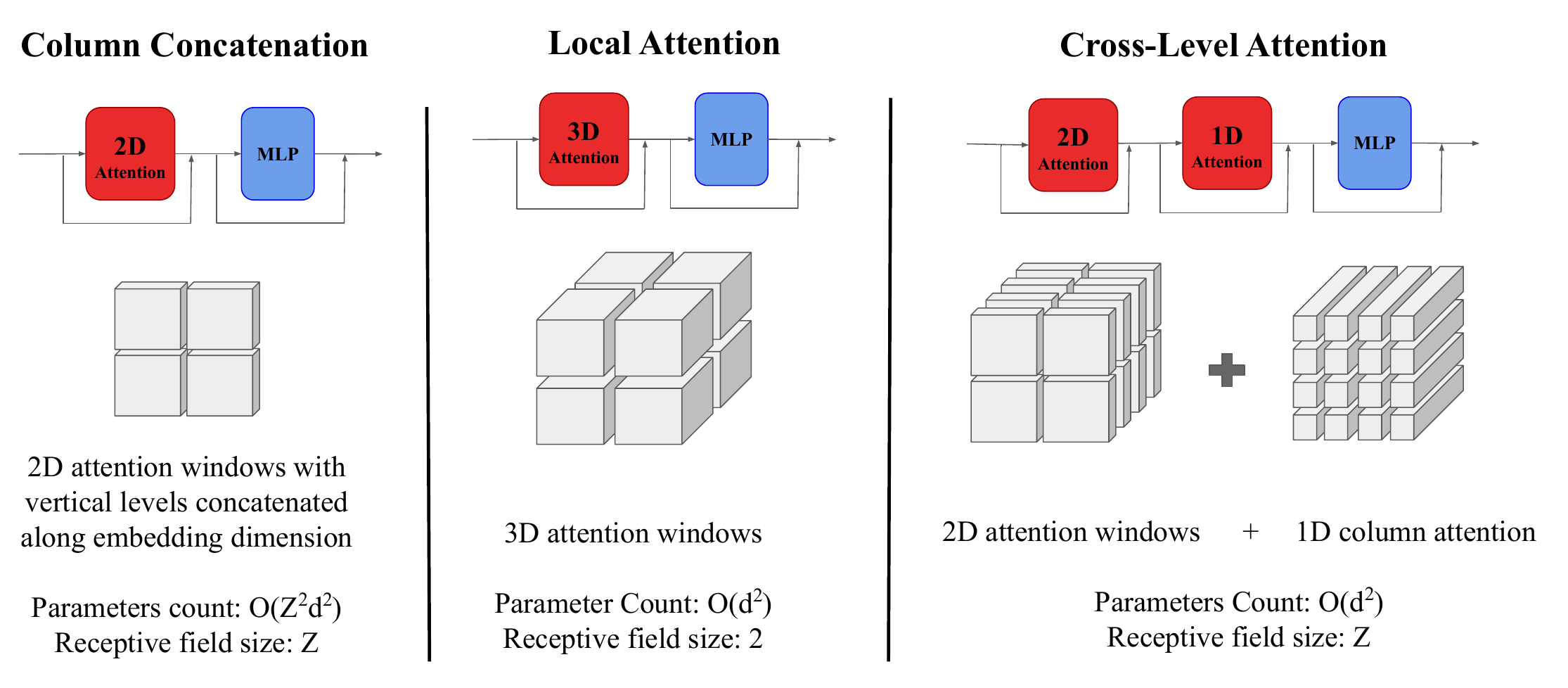}
    \caption{Visual comparison of attention schemes used in FuXi/Stormer (left), Pangu-Weather (middle) versus ours (right). Our design has the highest receptive field size without requiring a number of parameters scaling quadratically with respect to the number of layers $Z$.}
    \label{fig:attention}
\end{figure}

\paragraph{Improving attention with Cross-Level Attention (CLA).} We propose making all vertical features interact by adding a column-wise attention mechanism which we call Cross-Level Attention (CLA), that processes data along the vertical dimension of the tensor only. Considering the column data as a sequence of size $Z$, the number of parameters in this attention module is $O(d^2)$ and does not depend on $Z$. 
We also remove the vertical interaction from the original implementation by using horizontal attention windows of shape $(1, 6, 12)$, which reduces the cost of attention. The resulting attention scheme shares similarities with axial attention \cite{ho2019axial} with a decomposition of attention into two parts: column-wise attention and local horizontal 2D attention. See Figure \ref{fig:attention} for an illustration of our proposed attention scheme, compared to other attention methods. Axial attention has also been used in MetNet-3 \cite{andrychowicz2023deep} and SEEDS \cite{li2023seeds}.

\paragraph{Additional design choices.}
The standard for vision transformers used in dense tasks, like weather forecasting, is to use a strided deconvolution layer for the final projection \cite{bi2022pangu, chen2023fuxi}. However, this layer generally produces unphysical artifacts. We solve the issue by initializing the deconvolution layer with the ICNR scheme \cite{aitken2017checkerboard}. We also use the SwiGLU activation function \cite{shazeer2020glu} in the network's MLP layers instead of ReLU. SwiGLU is the de facto standard for modern Large Language Models (LLMs) and brings a significant improvement to the model's performance. Finally, following GraphCast, we provide additional information to the model (hour and month of desired forecast) with adaptive Layer Normalization \cite{perez2018film}.

\subsection{Training and evaluating deterministic weather models}

Deterministic ML models, which we note $f_\theta$ are trained by adjusting their parameters $\theta$ to minimize the weighted mean squared error (MSE) between the predicted state $f_\theta(\x_t)$ and the ground truth state $\x_{t+\delta}$:

\begin{equation}
    \label{deterministic_mse}
    \mathcal{L}(f_\theta(\x_t), \x_{t+\delta}) = \mathbb{E}_{t \in \mathcal D} \Vert f_\theta(\x_t) - \x_{t+\delta} \Vert_S^2.
\end{equation}

The norm $\Vert \cdot \Vert_S$ is a norm over weather states, which is a weighted mean of squared values in the tensor. The weighting comes from two sources: first, latitude weighting accounts for the area distortion introduced by the latitude/longitude representation of spherical data. Second, we use the same coefficients as GraphCast \cite{lam2022graphcast} to sum the contribution of different physical variables. In particular, we use a coefficient proportional to air density for upper-air variables and specific coefficients for surface variables: $0.1$ for mean sea level pressure and 10-meter wind components, and $1$ for 2-meter temperature. 2-meter temperature is assigned a much higher coefficient due to its importance as a predicted variable.

\begin{figure}[hb]
    \centering
    \includegraphics[width=0.8\linewidth]{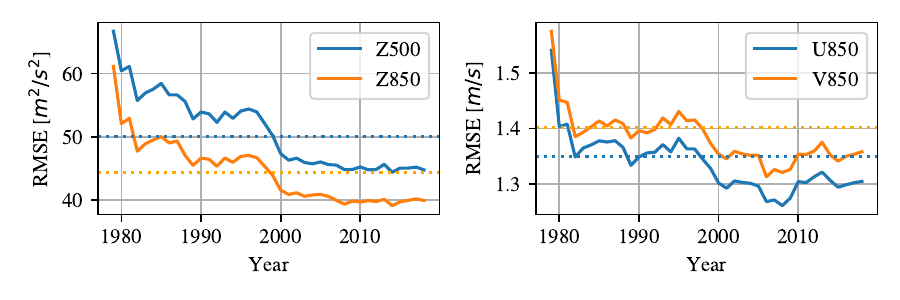}
    \caption{Geopotential (left) and wind speed (right) RMSE of a model without multi-step fine-tuning, for each year in the training set. The error is lower in the recent past, which we attribute to a more observed, constrained, and predictable dynamical system. Additionally, test RMSE (year 2020) are shown in dotted lines, which shows some overfitting compared to the scores in 2018 (last year in train set).}
    \label{fig:finetunefig}
\end{figure}

\paragraph{Detailed protocol for training ArchesWeather.} We train two versions of our architecture: ArchesWeather-S (16 transformer layers, 44M parameters) and ArchesWeather-M (32 layers, 84M parameters).
The training protocol is as follows:
\begin{itemize}
    \item \textbf{Phase 1}: Train models with the MSE loss for $250k$ steps.
    \item \textbf{Phase 2}: Fine-tune models on recent data (2007-2018) for $50k$ steps. We find that forecasting models have a higher error in the first half of the training period 1979-2018 (see Figure \ref{fig:finetunefig}, which we attribute to ERA5 being less constrained in the past due to a lack of observation data \cite{hersbach2020era5}. We found that fine-tuning on recent data helps to overcome this distribution shift, a process which we call \textit{recent past fine-tuning} or RPFT for short.
    \item \textbf{Phase 3 (optional)}: Fine-tuning on autoregressive rollouts for $20k$ steps. Models optimized for a single lead time of 6 to 24 hours are suboptimal in terms of RMSE scores when used autoregressively for longer lead times predictions. A standard procedure to improve RMSE scores at longer lead times is to fine-tune the model on auto-regressive rollouts, summing the MSE losses at each rollout step before backpropagation \cite{lam2022graphcast, chen2023fuxi, nguyen2023scaling}. We fine-tune our model on auto-regressive rollouts of length $2$ days for 8k steps, $3$ days for 8 steps and $4$ days for 4k steps, with gradients from the full trajectory, and call this procedure \textit{multi-step fine-tuning}.
\end{itemize}

Training the ArchesWeather-M model takes around 2.5 days on 2 A100 GPUs.

Finally, we train a small ensemble of our M models, called ArchesWeather-Mx4, by independently training multiple models with different random seeds and averaging their outputs at inference time. This helps remove modeling errors due to initialization and better approximates the true ensemble mean \cite{krogh1994neural}. Since a better ensemble mean prediction is desirable in residual modelling, we use ArchesWeather-Mx4 as our default underlying deterministic model in our generative training pipeline.

\paragraph{Limitations of deterministic models.} Due to the MSE loss, deterministic models produce an approximation of the mean of all possible future next states $\mathbb{E} [\x_{t+\delta} | \x_t]$, which is a weather state that is smooth and not physically consistent due to uncertainty. 

This raises several problems: first, computing reliable statistics of extreme events from forecasts requires a faithful reproduction of the true distribution; second, smooth forecasts tend to misrepresent the full weather variability, even when using initial condition perturbation to get a set of possible trajectories.
Third, the multi-step dynamics of ML weather models are not well represented. Indeed, ML models have never seen smooth states as input, yet they are expected to be used autoregressively as emulators. Empirically, autoregressive rollouts do work, but models are often fine-tuned with multi-step rollouts to improve RMSE scores at longer lead times, which increases smoothing even more due to minimizing MSE scores for a more distant and uncertain future. 

The trade-off between minimizing RMSE and producing physically consistent weather states \cite{bouallegue2024accuracy} implies that RMSE of individual forecasts is not the correct metric to train or evaluate weather models. One can only get the best of both worlds by producing a set of weather trajectories all physically consistent, and only expecting the mean of these predicted trajectories to reach the lowest possible mean squared error.

To address probabilistic weather generation, we use \textbf{flow matching models}, a modern variant of \textbf{diffusion models}.

\subsection{Training flow matching weather models}

\begin{figure}[ht]
    \centering
    \includegraphics[width=\linewidth]{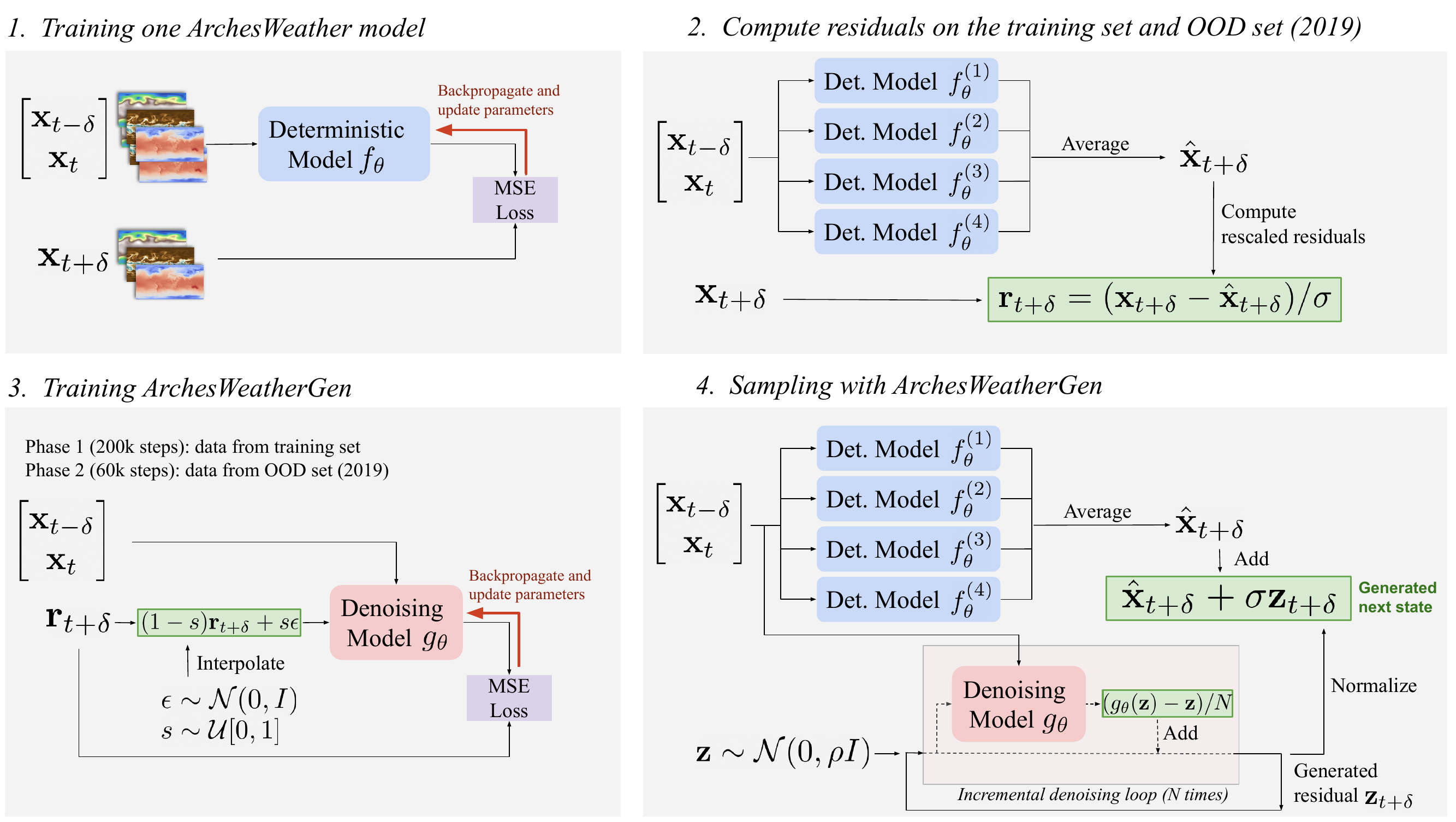}
    \caption{Main overview of our training pipeline of ArchesWeatherGen. (1) We train four ArchesWeather models by training neural networks to predict the next state with MSE loss. (2) We compute normalized residuals on the training and OOD sets, using ArchesWeather-Mx4 which is the average of 4 ArchesWeather models. (3) We train our flow matching model on the residual data $\r$, i.e we train a neural network to map residuals corrupted with gaussian noise to their uncorrupted version. (4) We sample ArchesWeatherGen by predicting the mean component with ArchesWeather-Mx4, then iteratively denoise gaussian noise to generate a residual sample, which is normalized and added to the mean to recover a complete sample of $\x_{t+\delta}$. The sampling process is then used autoregressively starting again with $t = t + \delta$ to generate multi-step trajectories. In the paper, we consider $\delta=24h$ and $\rho=1.05$.}
    \label{fig:main_diagram}
\end{figure}

We now discuss the background and motivation for training ArchesWeatherGen. Before going into the details, we present an overview of our methodology in Figure \ref{fig:main_diagram}.

Our goal is to sample weather trajectories given an initial condition by learning the transition distribution: we train a model to sample the distribution of next state $\x_{t+\delta}$ given the current state $\x_t$ and previous state  $\x_{t-\delta})$, and then compose our model autoregressively. This corresponds to the Markovian approximation:

\begin{equation}
p((\x_{t+k \delta})_{1\leq k \leq K} | \x_t, \x_{t-\delta}) \simeq \prod_{1\leq k \leq K} p(\x_{t+k\delta} | \x_{t+(k-1)\delta}, \x_{t+(k-2)\delta})
\end{equation}

In the remainder of the paper, we will write the transition function $p(\x_{t+\delta} | \x_t)$ and implicitly consider the dependency on $\x_{t-\delta}$ to simplify notations and presentation.

\paragraph{Diffusion models and flow-matching models.} 
For a general introduction to diffusion-based models (broadly defined), we refer the reader to \cite{sohl2015deep, ho2020denoising, nichol2021improved, weng2021diffusion, song2020denoising}. Here, we present concisely how these models are trained and used to model a transition distribution $p(\x_{t+\delta} | \x_t)$.
Diffusion-based models are trained to denoise samples from the data distribution that have been corrupted with Gaussian noise at all signal-to-noise ratios. For the original implementation of diffusion models, DDPMs, samples are mixed with Gaussian noise $\epsilon \sim \mathcal{N}(0, I)$ using the function $\z_{t+\delta}(s) = \sqrt{\alpha_s} \x_{t+\delta} + \sqrt{1-\alpha_s} \epsilon$ . Here, $s$ is the diffusion timestep, going from $0$ to $1$, which controls the level of noise added to the sample through the noise schedule $\alpha_s$. With $\alpha_0 = 1$, no noise is added and $\z_{t+\delta}(0)$ is a sample from the data distribution; with $\alpha_1 = 0$, the sample is completely corrupted and $\z_{s+\delta}(1)$ is a sample from the Gaussian distribution.
Given samples corrupted with all scales of noise levels, diffusion model training optimizes a neural network $g_\theta$ to predict the original samples using the loss function
\begin{equation}
\label{eq:denoising_diffusion}
    \mathcal{L} = \mathbb{E}_{x_t \in \mathcal{D}, s\sim \mathcal{U}[0, 1]} \; \gamma_s \Vert g_\theta(\x_t, \z_{t+\delta}(s)) - \x_{t+\delta } \Vert^2_S, 
\end{equation}

where $\x_t$ is given as input to the diffusion model to model the conditional distribution $p(\x_{t+\delta} | \x_t)$ and not just the unconditional distribution $p(\x_{t+\delta})$. 

The coefficients $\gamma_s$ control the relative contributions of the denoising objective at each noise level $\alpha_s$. Another common training objective is to predict the noise $\epsilon$ added to samples \cite{ho2020denoising}, which is equivalent to predicting samples (since we can find the original sample $\x_{t+\delta}$ from the predicted noise $\epsilon$), up to a change of coefficients $\gamma_s$. While sample prediction is easy at high signal-to-noise ratios (SNR) and difficult at low SNR, noise prediction is the exact opposite: easy at high SNR and hard at low SNR. To rebalance the difficulty of the learning objective across all noise levels, we follow the coefficients $\gamma_s = \sqrt{SNR(s)} = (1-s)/s$ in \cite{yu2023debias} that provide a more stable loss.

Flow matching models \cite{lipman2022flow} are a variant of diffusion models trained with a linear interpolating scheme for the noising function: $\z_{t+\delta}(s) = (1 - s) \x_{t+\delta} + s \epsilon$ for $s \in [0, 1]$. They are usually trained to predict $\x_{t+\delta} - \epsilon$ instead of $\x_{t+\delta}$, which corresponds to the flow velocity between the noise $\epsilon$ and the sample $\x_{t+\delta}$. However, we can also train them as denoising models, and get back the original flow target by interpolating the model output with $\z_{t+\delta}$.  Finally, the steps $s$ can be sampled nonuniformly, e.g., by using the sigmoid of a normal distribution as done in Stable Diffusion 3 \cite{esser2024scaling}, which we also use in our method.

\paragraph{Sampling diffusion-based models.} For both DDPMs and flow matching models, samples can be obtained by starting with a variable $z \sim \mathcal{N}(0, I)$, and iteratively moving it towards the data distribution by using the denoising model as target, giving rise to a neural ODE \cite{dhariwal2021diffusion, couairon2022diffedit}. For flow matching models, the ODE is $d\z_s = (g_\theta(\z_s) - \z_s) ds$, where $g_\theta$ is the trained denoiser. The simplest and most common choice to solve that ODE is an Euler scheme with the initial condition $z(0) \sim \mathcal{N}(0, I)$ and a constant step size, which is the one we use in this paper. The number of discretization steps used in that process can be freely chosen, which we set to 25 based on initial experiments.

\paragraph{Learning diffusion-based models on spatio-temporal data.} Since the underlying physical model used to produce the ERA5 dataset is deterministic, aleatoric uncertainty is driven mainly by the data assimilation process and the equirectangular sampling at 1.5º instead of the higher resolution natural representation of the dataset. At a lead time of $24h$,  the uncertainty is rather small, as demonstrated by the very good skill that deterministic ML models have on this time scale. As a result, the conditional distribution $p(\x_{t+\delta} | \x_t)$ is narrowly centered around its expectation $\mathbb{E} [\x_{t+\delta} |\x_t]$. Therefore, the states can be decomposed as
\begin{equation}
    \x_{t+\delta} = \mathbb{E}[\x_{t+\delta} | \x_t] + \r_{t+\delta},
\end{equation}
where $\r_{t+\delta}$ is a centered residual ($\mathbb{E}(\r_{t+\delta} | \x_t) = 0$) with a relatively small variance compared to the unconditional variance of states $(\x_t)_{t \in \mathcal D}$.

Training a deterministic model $f_\theta$ with MSE loss approximates the expectation term $\mathbb{E} [\x_{t+\delta} |\x_t]$, because it is the optimal minimizer of the MSE objective:
\begin{equation}
\label{eq:expected_deviation}
    \mathbb{E} [\x_{t+\delta} | \x_t] =  \text{argmin}_M \;\; \mathbb{E}_{\x_{t+\delta} \sim p(\x_{t+\delta} | \x_t)}\Vert M - \x_{t+\delta} \Vert ^2.
\end{equation}
This means that we can use any deterministic model $f_\theta$ trained with MSE loss to approximate the expectation term. Then, given a choice of $f_\theta$, we train a flow matching model $g_\theta$ to model residual data $\r_{t+\delta} = (\x_{t+\delta} - f_\theta(\x_t))/\sigma$ renormalized to unit variance, using the denoising loss function

\begin{equation}
\label{eq:residual denoising_diffusion}
    \mathcal{L} = \mathbb{E}_{x_t \in \mathcal{D}, s\sim \mathcal{U}[0, 1], \epsilon \sim \mathcal{N}(0, I)} \; \gamma_s \Vert g_\theta(\x_t, (1-s)\r_{t+\delta} + s\epsilon, f_\theta(
    \x_t)) - \r_{t+\delta}\Vert^2_S. 
\end{equation}

Finally, ensemble predictions are generated by sampling the generative model and adding it to the deterministic prediction $f_\theta(\x_t)$. Forecasts at longer lead times are obtained by repeating this process auto-regressively.

The advantages of this decomposition are threefold: (i) reduce the computational complexity of training weather generative models by factorizing the common work between trainings, (ii) leverage state-of-the-art deterministic models trained with a very high computational budget, and cheaply adapt them to a generative setting, (iii) the residual data have a simpler structure that can translate to a diffusion sampling ODE with straighter paths, which generally results in better sample quality with fewer sampling steps.
In the remainder of the paper, we call \textbf{residual data} or \textbf{residuals} the data where the output of a deterministic model has been removed and refer to this deterministic model as the \textbf{ underlying deterministic model}. We will call \textbf{residual generative model} the model that is trained on residuals, as opposed to the \textbf{generative model} that makes complete forecasts by summing outputs from the deterministic and residual generative model.

\paragraph{Overcoming difficulties arising from residual modeling.}\label{overfitting_underdispersion}

In our experiments, We have observed overfitting of our deterministic models, with a validation and test error a few percent higher than the training error (visible on Figure \ref{fig:finetunefig}). As a result, the distribution of residuals is different on the test set compared to the training set.
A manifestation of this is that residuals have a higher norm on the test set compared to the training set, because the norm of residuals corresponds to the error of the deterministic model used to compute them. Therefore, the generative model has to produce residuals at test time from a slightly different distribution compared to what it was trained on. 

The main problem that arises from this overfitting issue is under-dispersion. Our residual generative model learns to model centered data on the training set, which means that the ensemble mean of residuals will be 0. Therefore, the ensemble mean matches the prediction of the underlying deterministic model. Consequently, the Ensemble Mean RMSE corresponds to the error of the underlying deterministic model, which is higher on the test set than on the training set. The Ensemble Mean RMSE being the discriminator of the spread-skill ratio \ref{sec:evaluation}, this results in spread-skill ratios below one, hence underdispersion.

To mitigate this issue, we introduce \textbf{OOD fine-tuning} (for \textbf{O}ut-\textbf{O}f-\textbf{D}istribution), where we fine-tune the generative model on data that was not used for training the deterministic model. Residuals coming from this data source have a distribution closer to that which should be modeled on the test set. 
We also introduce \textbf{noise scaling}, which consists of scaling the initial noise for the sampling process, with a coefficient slightly higher than 1. The generative model then projects this noise distribution of a slightly higher variance, to an output distribution that also has a slightly higher variance, better matching the expected dispersion of members. In the experiments, we use a noise scaling coefficient of $1.05$, which roughly corresponds to the percentage of overfitting observed for our underlying deterministic models.

\paragraph{Protocol for training ArchesWeatherGen.}

To sum up our training protocol for ArchesWeatherGen, we first choose a base deterministic model $f_\theta$, which approximates the conditional mean $\mathbb{E}[\x_{t+\delta} | \x_t]$.
By default, we use an ensemble of 4 ArchesWeather-M models. Then, we train a flow matching model on the renormalized residual values $\r_{t+\delta} = (\x_{t+\delta} - 
f_\theta(\x_t))/\sigma$, in two phases:
\begin{itemize}
    \item \textbf{Phase 1}: Train the denoising model on 1979-2018 data for $200k$ steps.
    \item \textbf{Phase 2 (OOD fine-tuning)}: Fine-tune on 2019 data for $60k$ steps, to adapt the generative model to residuals of higher norm corresponding to data not overfitted by the underlying deterministic model.
\end{itemize}

Because this flow matching model is used in a neural ODE at inference time, generating a single sample requires many calls to the neural network (25 in our case). That is why we chose the ArchesWeather-S architecture to further reduce inference costs.

\subsection{Evaluation.}\label{sec:evaluation} 

We evaluate deterministic models with the latitude-weighted Root Mean Square Error (RMSE). For generative models, we use CRPS, Ensemble Mean RMSE, Energy Score, Brier Score (with a 1\% threshold) and spread-skill ratio. Because of a statistical bias in the estimation of these scores due to a finite number of ensemble members, we use a "fair" version of these metrics by removing this bias, which corresponds to their theoretical value in the limit of an infinite ensemble. This allows us to compare scores for ensembling methods with different ensemble sizes. We indicate the fair version of these metrics by prefixing them with the letter f, e.g. fCRPS for the fair CRPS \cite{zamo2018estimation}.

The \textit{Ensemble Mean RMSE} is the RMSE of the ensemble mean, which captures how well generated members capture the expectation of the distribution. Unlike in single-member prediction, there is no trade-off between this metric and smoothness: an ensemble mean can display the correct smoothness (reflecting uncertainty) with members that are all physically consistent.

The \textit{Continuous Ranked Probability Score} \cite{gneiting2007strictly} is a widely used metric for probabilistic forecasts, which allows us to assess whether the pointwise marginal distributions predicted by a model correctly match the ground-truth marginal distributions. Given that we are interested in the conditional probability distribution given the current state $\x_t$, there is only one sample of the transition distribution (the following state $\x_{t+\delta}$ in ERA5), but we can still compute the CRPS with this unique sample and average over all initial conditions in the test set. For the same lead time as training (24h in our case), the CRPS only evaluates whether the marginal distributions are correct, and could be optimized without producing a physically consistent prediction. However, for ML models that are used in an autoregressive manner, reproducing the correct marginals at longer lead times requires producing physically consistent predictions at intermediate lead times, otherwise the models are run on out-of-distribution inputs which degrade the trajectories. Hence, CRPS it is a representative metric of the quality of model outputs, beyond marginal distributions. 

\textit{Brier scores} \cite{roulston2007performance} are a measure of the forecasting skill for extreme events. Given a quantile threshold $q$, they measure whether the empirical probability of a physical variable exceeding a threshold $q$ is close to the ground truth (1 if the threshold was exceeded in ERA5, 0 otherwise). In this paper, we compute Brier scores using thresholds of $<1\%$ and $>99\%$ quantiles of the 1990-2018 climatology, and then averaged together to get a unique score for the $1\%$ tails of the distribution.

\textit{Spread-skill ratio}  is a measure of the dispersion of an ensemble. It assesses whether the predicted standard deviation of members matches the Ensemble Mean RMSE, which should be the case for a perfect prediction \cite{fortin2014should}. The model is underdispersive with a spread-skill ratio of less than 1 and overdispersive with a spread-skill ratio greater than 1, assuming relatively small forecast bias \cite{hamill2001interpretation}.

\paragraph{Summary metrics.} For all metrics, latitude re-weighting is used to account for the sampling distortion introduced by the equirectangular projection, so that the contribution of each pixel is proportional to its area on the sphere. 
We focus on evaluating key \textit{headline} variables: Z500, Q700, T850, U850, V850, T2m, SP, U10m, and V10m. We also use skill scores to compare metrics to a reference model, allowing us to assess relative performance with respect to this model. For skill scores, higher is better: an RMSE skill score of $5\%$ means that the evaluated model gets an RMSE score $5\%$ lower compared to the reference model. Skill scores have the additional benefit to be dimensionless, and therefore can be averaged, a property that we use to get a representative score across all variables. For instance with RMSE, we define the global RMSE skill score as the average 
\begin{equation}
     \text{RMSE-ss}(model) =  \frac{1}{\Vert \mathcal V \Vert} \sum_v \Big(1 - \frac{\text{RMSE}_v(model)}{\text{RMSE}_v(ref)}\Big)
\end{equation}

where $RMSE_v(X)$ is the RMSE of model $X$ on physical variable $v$ and $ref$ is the reference model. We use the IFS models as reference, with HRES for evaluating deterministic models and ENS for evaluating probabilistic models. In all cases, these reference models are evaluated using their own analysis data as ground truth instead of ERA5, following WeatherBench \cite{rasp2023weatherbench}.

\section{Experiments}

We first evaluate our deterministic model ArchesWeather with RMSE, at a lead time of 24h as well as longer lead times through auto-regressive rollouts. We then showcase some limitations of deterministic models and evaluate ArchesWeatherGen, with the ensemble metrics defined above. Finally, we validate some of our design choices in section \ref{section:gen_ablation}.

\subsection{Evaluating ArchesWeather deterministic models}

\subsubsection{Main results at 24h lead time}

Table \ref{full_results} shows RMSE scores of ArchesWeather compared to state-of-the-art ML weather models, including Pangu-Weather and GraphCast, SphericalCNN \cite{esteves2023scaling}, NeuralGCM at 1.4º (50 members ensemble), and Stormer.
Data is from WeatherBench 2, except Stormer where we evaluated outputs provided by the authors. Some models do not include surface variables and as such their performance cannot be reported.

The ArchesWeather-M base model largely surpasses the SphericalCNN model for upper-air variables, with a training budget of around 9 V100-days, i.e., 40 times smaller. At 24h lead time, the ArchesWeather-Mx4 ensemble version outperforms the 1.4º NeuralGCM ensemble (50 members) on upper-air variables. They perform on par with the original Pangu-Weather(0.25º) and Stormer(1.4º), except for wind variables (U850, V850, U10, V10) where notably Stormer is consistently better. This might be due to the higher training budget (256 V100-days), bigger models, or averaging outputs from 16 model forward passes (more details in Appendix \ref{app:sota}). Investigating this discrepancy is left for future work. 

\input{figs/weatherbench_deterministic}

\subsubsection{Multi-step evaluation}\label{app:multi-step}

We now report scores of deterministic models evaluated at longer lead times through autoregressive rollouts, both with multi-step fine-tuning and without.

\begin{figure}[t]
    \centering
    \includegraphics[width=\linewidth]{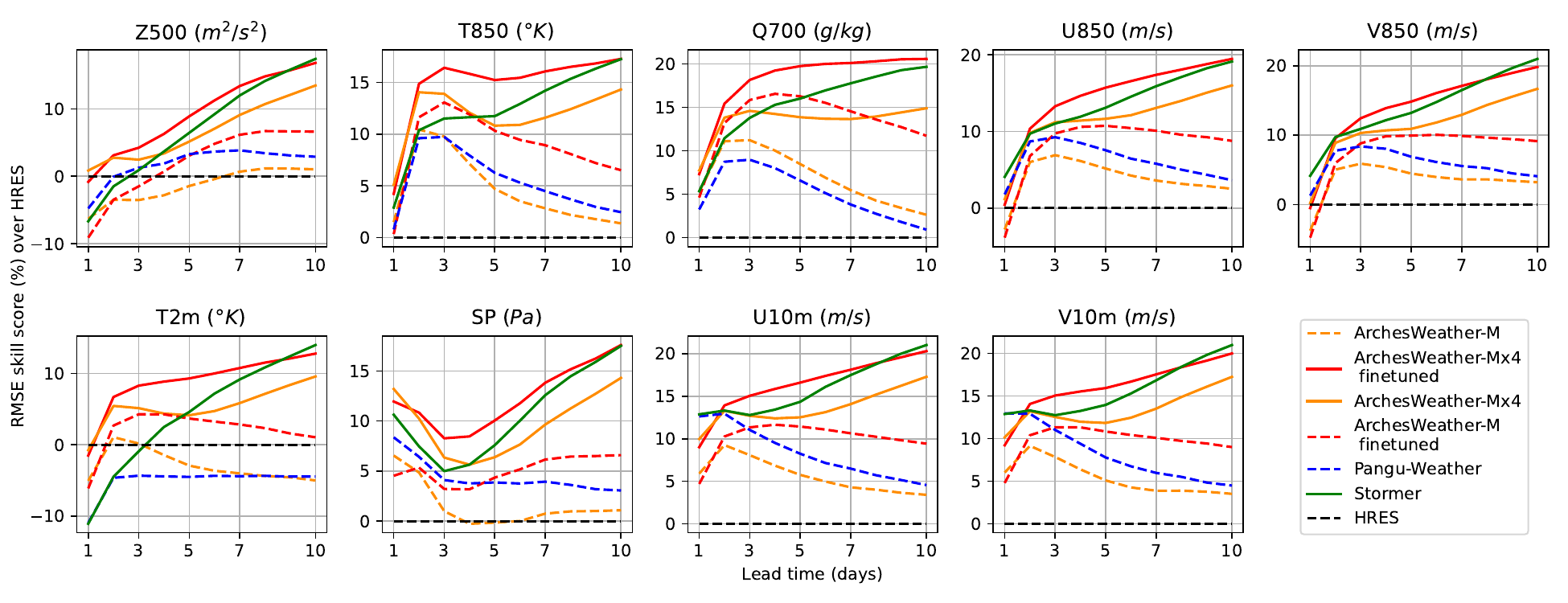}
    \caption{RMSE skill scores of weather models for lead times up to 10 days. Models that don't use ensembling are shown in dotted lines. We can see that ArchesWeather-Mx4 surpasses Stormer on all headline metrics for lead times up to 8 days, despite using 4 times fewer ensemble members (4 vs 16).}
    \label{fig:multi-step}
\end{figure}

In Figure \ref{fig:multi-step}, we plot the RMSE skill scores (with respect to HRES) of our models ArchesWeather-M (without multi-step fine-tuning), ArchesWeather-Mx4 (ensemble of 4 models without multi-step fine-tuning), and ArchesWeather-Mx4 fine-tuned (ensemble of 4 models, each with multi-step fine-tuning). The competing methods that we use for reference are Pangu-Weather at 0.25º (closest to our model), and Stormer at 1.4º (best competing method at a similar resolution).

First, we see that the simple strategy of training 4 independent ArchesWeather models and averaging them allows us to reach better scores than GraphCast after 7 days, showing the importance of not relying only on multi-step fine-tuning to get good RMSE scores at longer lead times. 
Second, our ensemble of 4 fine-tuned models surpasses Stormer on all headline variables for lead times of 2 to 8 days. Interestingly, the better 24h RMSE scores for Stormer on wind variables (U850, V850, U10m, V10m) do not translate to better multi-step trajectories.

Multi-step fine-tuning on autoregressive rollouts allows one to improve the RMSE scores of single trajectories at longer lead times. For ensemble forecasting, only the ensemble mean should reach the minimum RMSE error, and not each member of the ensemble, meaning that multi-step fine-tuning is not needed. In the remainder of the paper, we use 24h-ahead underlying deterministic models without multi-step fine-tuning.

\subsubsection{Ablations of the deterministic model}

\input{figs/ablation_table}

Our main ablation experiment for deterministic models is presented in Table \ref{abl}, where we compare models without multi-step fine-tuning.
For a fair comparison between models, we decrease the embedding dimension (by about 5\%) when using CLA, so that all types of model have roughly the same parameter count. Compared to Pangu-Weather (retrained in the same setting as us), our model without Cross-Level Attention or fine-tuning on recent data (RPFT) still largely improves performance.
Adding our proposed Cross-Level Attention scheme significantly improves RMSE scores, reducing by half the RMSE difference with the IFS HRES. ArchesWeather with 16 layers reaches a lower error than using 32 layers without CLA (e.g., Z500 RMSE of 50.6 vs. 51.8). Fine-tuning the model on recent data for the last 50k steps brings some small additional benefit. Given that we relaxed the inductive bias for the 3D atmospheric interaction, we also asked whether it could be beneficial to remove the 3D prior entirely by flattening the data along the vertical dimension. This is the setting of the "2D ArchesWeather" in Table \ref{abl}, whose performance is largely below the best ArchesWeather models despite a much higher parameter count ($\sim$850M). This tends to show that the 3D inductive prior is still useful, and the factorized attention better generalizes compared to fully connected matrices along the vertical dimension.

\subsection{Comparing generative models to deterministic models}

\begin{figure}[ht]
    \centering
    \includegraphics[width=\linewidth]{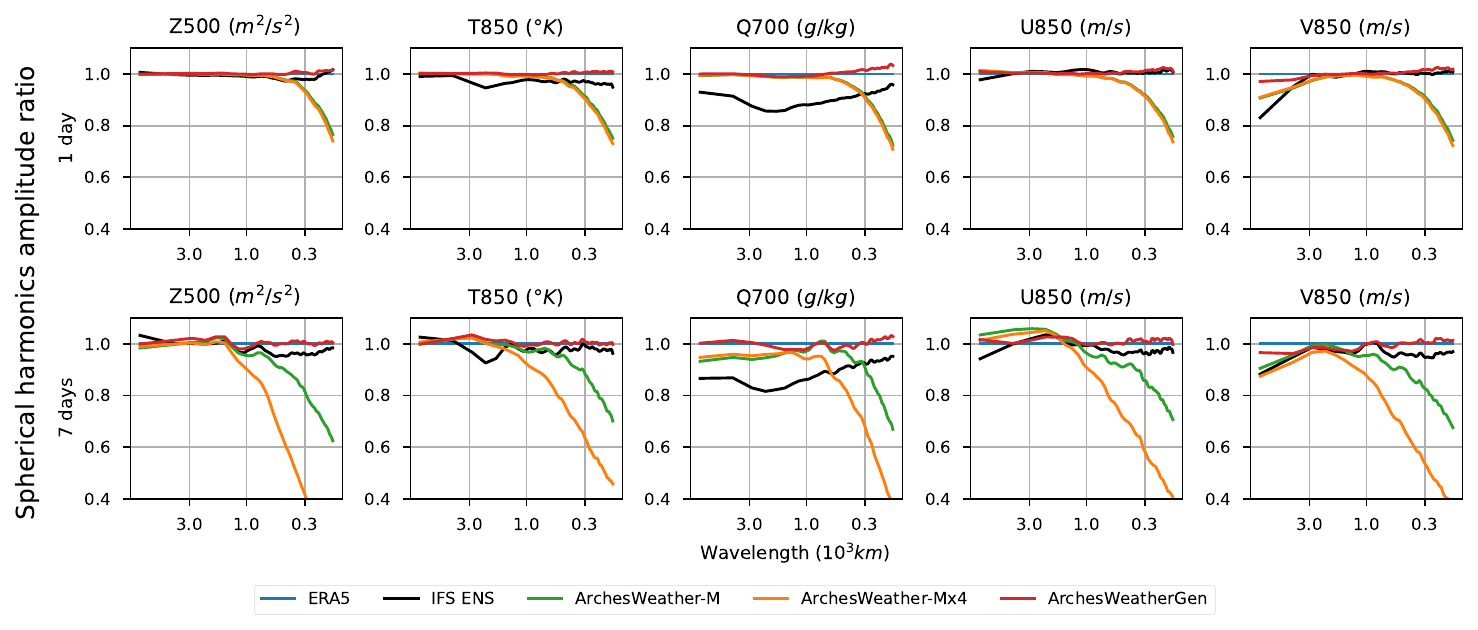}
    \caption{Spherical harmonics amplitude ratio for different models. For each model and wavelength, the energy at that wavelength is averaged across samples and divided by the corresponding averaged energy in ERA5. Our model's power spectrum is much  closer to ERA5's, and on average better than that of IFS ENS.}
    \label{fig:spectrum}
\end{figure}

\paragraph{Smoothness evaluation.} An important criterion for assessing the relevance of weather models is the level of physical consistency of predicted trajectories. We have mentioned that deterministic models predict smooth outputs, a phenomenon that can be measured with two metrics: activity, which is the standard deviation of the climatology-removed forecast across spatial locations, and power spectra. In Figure \ref{fig:spectrum}, we measure fidelity with respect to the power spectra of ERA5, by averaging the energy of each wavelength across generated forecasts and dividing by the averaged energy at that wavelength in ERA5. We can see that the deterministic models do not have enough energy at short spatial scales, due to smoothing. ArchesWeatherGen's power spectra are much closer to ERA5's. It is also better than IFS ENS on Z500, T850 and Q700 at a 1 day lead time, and better on all variables at a 7-day lead time.

At a one-day lead time, we can see that our model has very slightly more energy in the short wavelengths compared to ERA5, especially on Q700. However, this error does not accumulate over time, and the 7-day power spectrum ratios are of similar quality compared to the one day ratios.

\begin{figure}[ht]
    \centering
    \includegraphics[width=\linewidth]{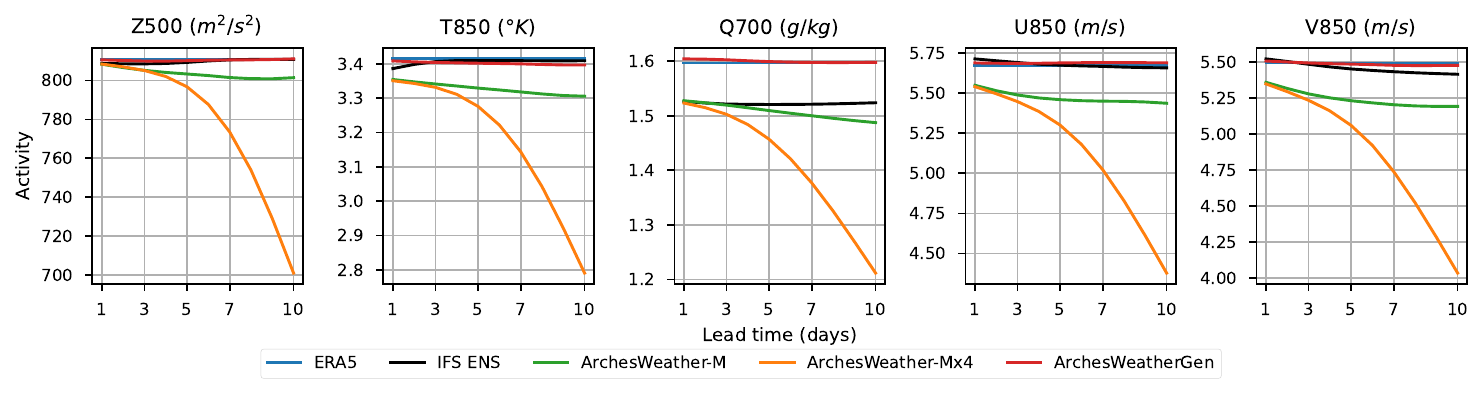}
    \caption{Activity (standard deviation of the climatology-removed forecast across spatial locations) of our models for lead times up to 10 days. ArchesWeatherGen's activity is very close to the ERA5 reference.}
    \label{fig:activity}
\end{figure}

Figure \ref{fig:activity} shows that the activity of our ArchesWeatherGen is much closer to the ground truth compared to our deterministic models. We can see the activity of deterministic models decreasing as the lead time increases, which shows that smoothing is stronger at longer lead times. This effect is particularly strong for the ArchesWeather-Mx4 ensemble, because
averaging the outputs of different models increases smoothing even more compared to a single RMSE-trained model.

\begin{figure}[ht]
    \centering
    \includegraphics[width=\linewidth]{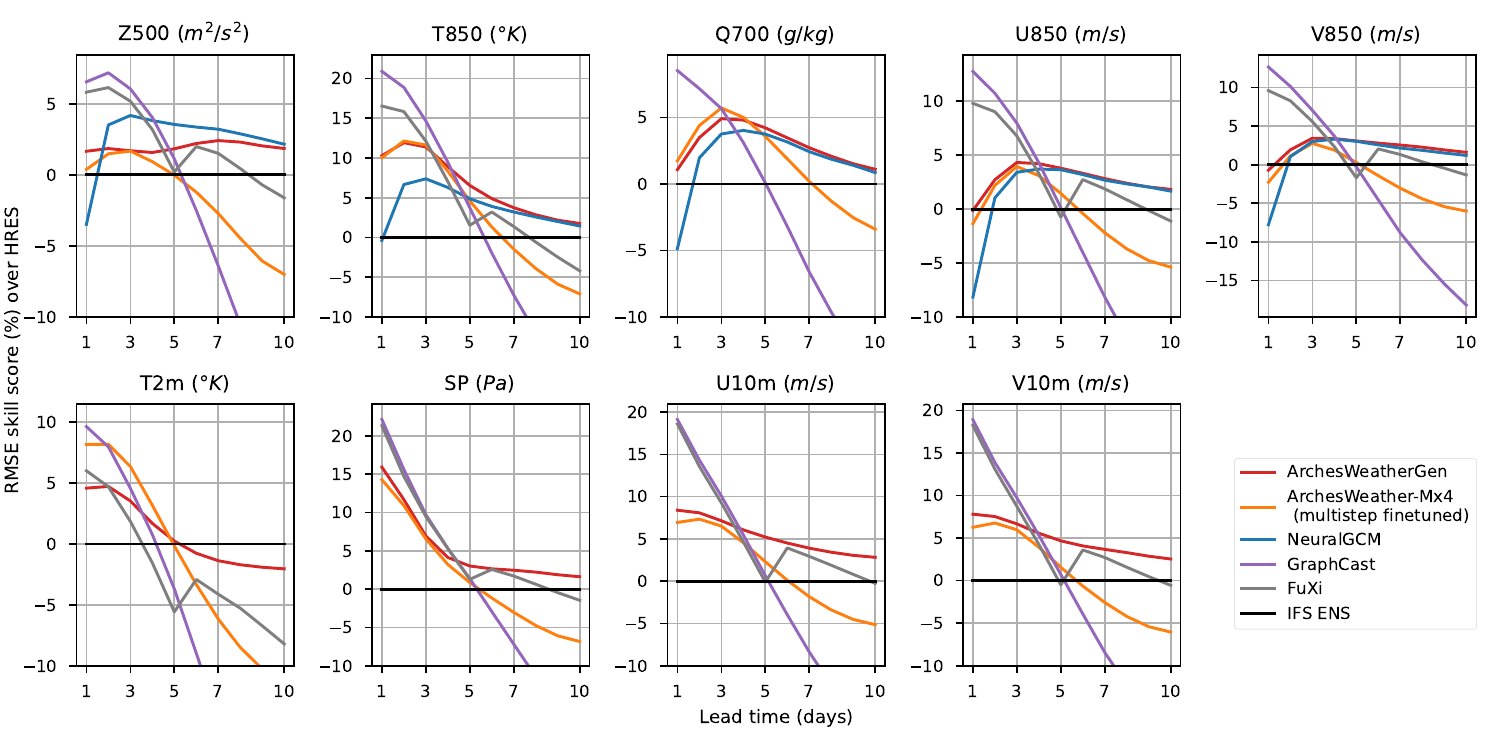}
    \caption{Ensemble Mean RMSE skill scores of ArchesWeatherGen, compared to our small deterministic ensemble ArchesWeather-Mx4, NeuralGCM, and state-of-the-art deterministic models (FuXi, GraphCast). ArchesWeatherGen reaches the best performance on all variables and lead times, except Z500, for lead times of 2 to 7 days.}
    \label{fig:rmse_ens}
\end{figure}

\paragraph{Better approximation of the ensemble mean with generative models.}
In this section, we evaluate Ensemble Mean RMSE skill scores of our ArchesWeatherGen model, compared to NeuralGCM, our small deterministic ensemble ArchesWeather-Mx4, and other state-of-the-art deterministic models (FuXi, GraphCast). Although Ensemble Mean RMSE does not directly measure the ability of models to predict the true distribution over possible futures, it is useful to measure whether predicted trajectories have the correct expectation or whether they are biased.
In Figure \ref{fig:rmse_ens}, we report the Ensemble Mean RMSE of our models compared to competing methods, including deterministic models. In this plot only, we do not correct for statistical bias for a more fair comparison between ensemble and deterministic models, since members generated with deterministic methods cannot necessarily be scaled easily (for instance, our ArchesWeather-Mx4 model would require to retrain models with other random seeds to generate more than 4 members). Using this non-corrected metric does favor methods that are able to produce many different ensemble members, which is a desired behavior.

Our ensemble of 4 deterministic models ArchesWeather-Mx4 has better scores than GraphCast, which shows the importance of using multiple members to approximate the ensemble mean. FuXi reaches better scores because it was trained to directly approximate the ensemble mean after 5 days, unlike GraphCast or ArchesWeather with multi-step fine-tuning. Ensemble methods (NeuralGCM and ArchesWeatherGen) better capture the Ensemble Mean compared to FuXi, which shows the advantage of using sampling ensemble members rather than predicting the ensemble mean with a neural network. Finally, ArchesWeatherGen reaches better Ensemble Mean RMSE scores than NeuralGCM for upper-air variables (NeuralGCM does not predict surface variables), except for geopotential. We make the hypothesis that this could be due to the nature of the Z500 field, which is smoother and likely very well modeled by NeuralGCM's dynamical core.

\subsection{Quantitative evaluation of generative models}

In this section, we quantitatively evaluate our main model ArchesWeatherGen, compared to other weather ensemble models: IFS ENS, NeuralGCM (which is the only ML-based ensemble model available in WeatherBench), and a baseline version called ArchesWeather-DDPM, which uses our neural network architecture as the backbone for DDPM diffusion models, without flow-matching, OOD fine-tuning, or noise scaling.

\begin{figure}[ht]
    \centering
    \includegraphics[width=\linewidth]{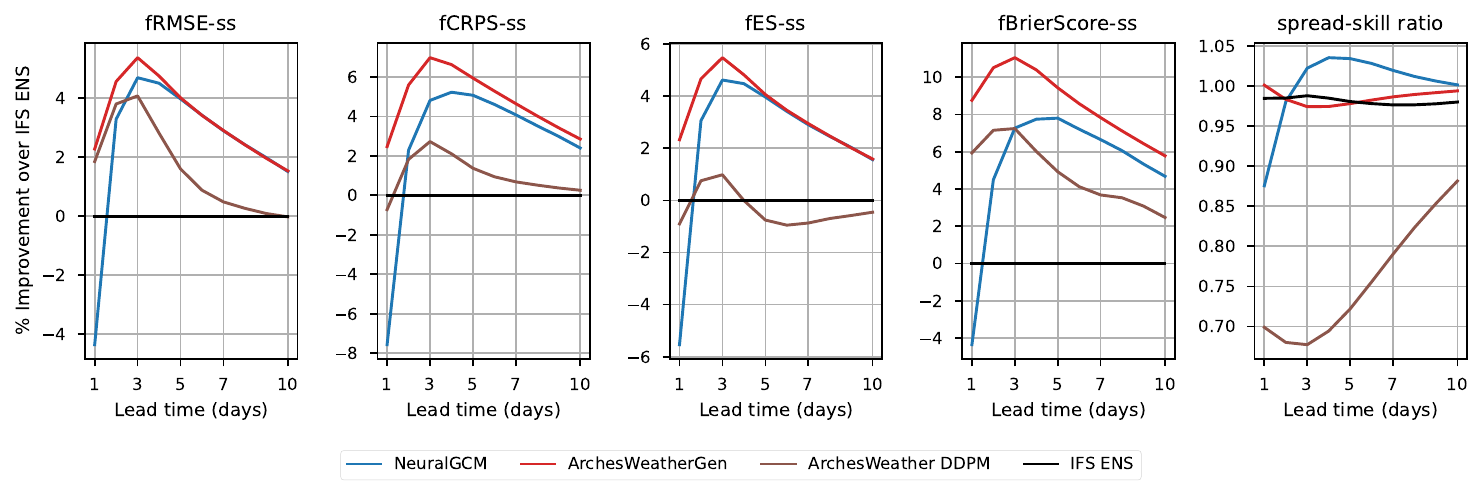}
    \caption{Evaluation of ArchesWeatherGen compared to other weather ensemble models. The scores are shown as relative improvement with respect to IFS ENS, for RMSE, CRPS, Energy score and Brier Score (1\% tails), and averaged over our headline upper-air variables: Z500, T850, Q700, U850, V850.
    ArchesWeatherGen reaches the best averaged metrics across all lead times, and has similar spread-skill ratio as IFS ENS.}
    \label{fig:main_gen}
\end{figure}

\paragraph{Summary of metrics.} In Figure \ref{fig:main_gen}, comparison between the different models is based on the key metrics for ensemble evaluation: Ensemble Mean RMSE (RMSE for short), CRPS, Energy Score (ES), Brier Score (average of scores with thresholds of 1\% and 99\%) and spread-skill ratio. All metrics are shown as relative improvement over IFS ENS. To be able to compare our averaged skill scores to NeuralGCM, on Figure \ref{fig:main_gen} only, we average skill scores only on the upper air headline variables of WeatherBench (Z500, Q700, T850, U850, V850).

We observe that ArchesWeatherGen has better scores than IFS ENS and even NeuralGCM for all lead times up to 10 days. Our model reaches Ensemble Mean RMSE scores similar to those of NeuralGCM for lead times greater than 5 days, which we had also observed in Figure \ref{fig:rmse_ens}.

ArchesWeatherGen is very slightly underdispersive, similar to IFS ENS. Compared to ArchesWeather-DDPM, we first see that our model brings an important improvement over the original DDPM diffusion models, with relative CRPS scores roughly 4\% better for all lead times. The DDPM baseline is very underdispersive, with a spread-skill ratio of $0.7$ for 24h, increasing to $0.88$ at a 10-day lead time. which flow matching and our other proposed improvements address.

\begin{figure}[t]
    \centering
    \includegraphics[width=\linewidth]{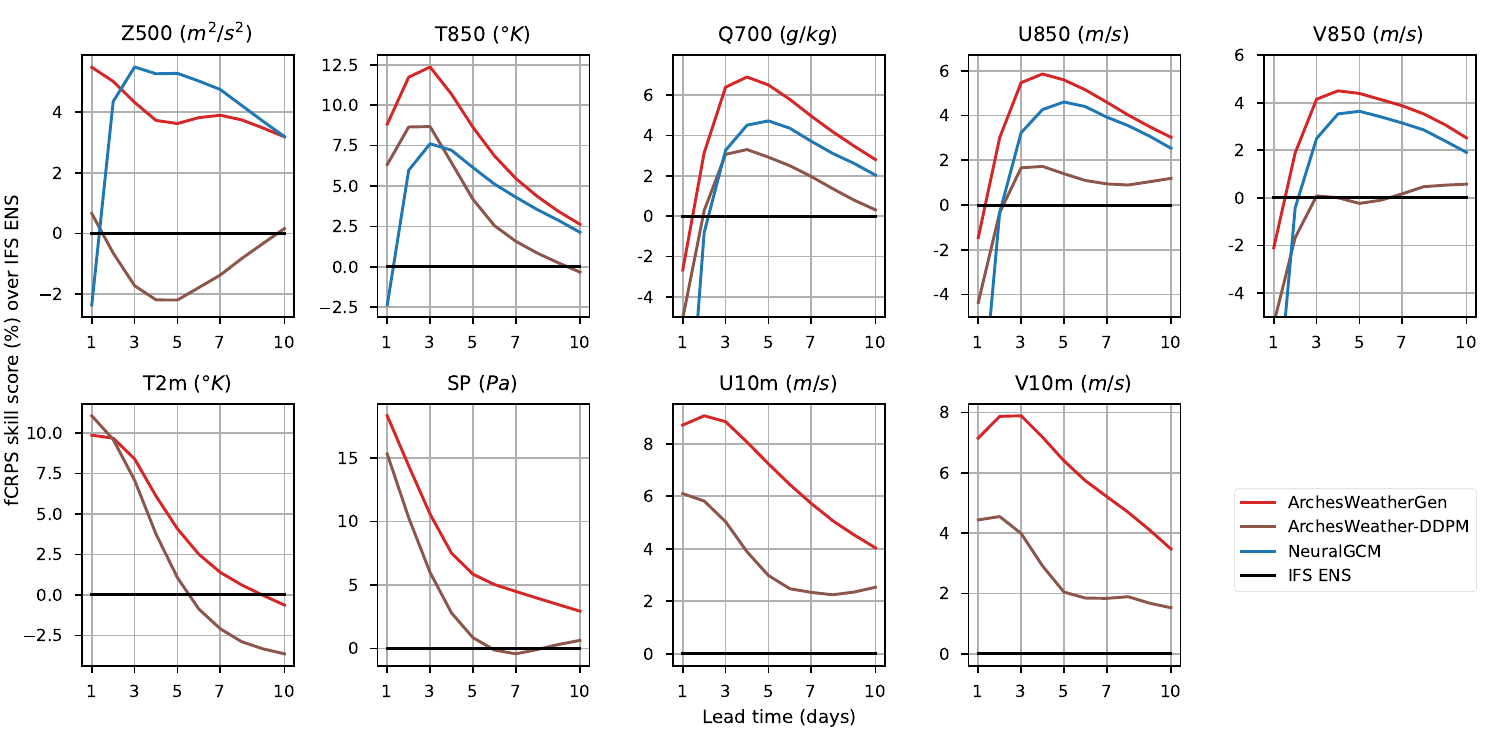}
    \caption{fCRPS skill scores for headline physical variables in WeatherBench, comparing our models to NeuralGCM and IFS ENS. ArchesWeatherGen surpasses IFS ENS, ArchesWeather-DDPM, and has better scores than NeuralGCM, except on geopoential, between 3 and 10 days.}
    \label{fig:multistep_crps_main}
\end{figure}

\paragraph{Per-variable CRPS skill scores.} 
While in the last section we have shown metrics averaged over headline variables, we now investigate scores for each physical variable independently. We focus here on the CRPS scores, that measure the quality of marginal distributions as well as requiring the auto-regressive models to make physically consistent predictions to have good scores at longer lead times. We report these scores across lead times in Figure \ref{fig:multistep_crps_main}, comparing ArchesWeatherGen with ArchesWeather-DDPM and NeuralGCM.

ArchesWeatherGen reaches better CRPS skill scores than IFS ENS across all headline variables and lead times, on average by 5.3\%. It is slightly worse only for T2m at lead times longer than 9 days. It improves upon NeuralGCM on all headline variables and lead times, except geopotential for lead times of 3 to 10 days. This advantage for NeuralGCM, which we have already observed on Ensemble Mean RMSE, might again be due to the dynamical core modeling, which is better suited for capturing the evolution of geopotential height.
Compared to DDPM, using our model yields significant CRPS improvements on all variables, generally by 2-4\% relative points across lead times, and up to 5-6\% for geopotential.
As geopotential is the variable for which overfitting of the underlying deterministic model is the strongest, we believe that this larger gain for geopotential demonstrates that our proposed improvements to tackle overfitting are working.

\paragraph{Reliability of ensemble forecasts.}

\begin{figure}[ht]
    \centering
    \includegraphics[width=\linewidth]{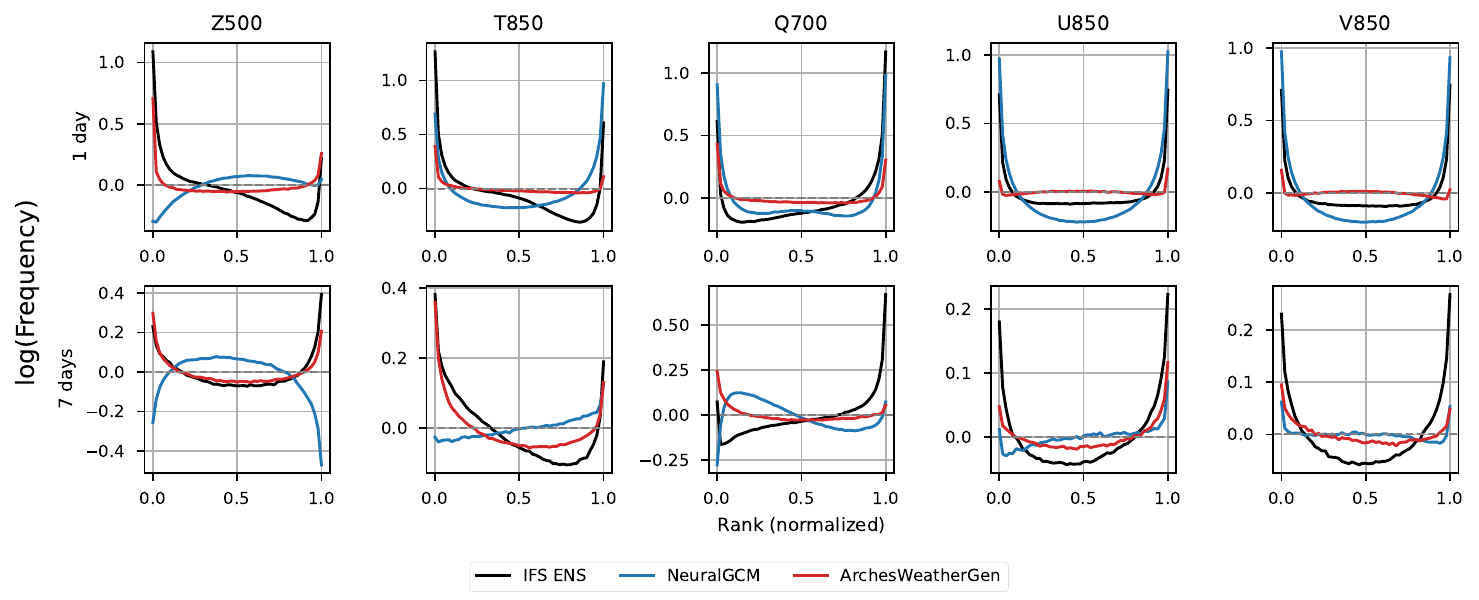}
    \caption{Rank Histogram of ArchesWeatherGen compared to NeuralGCM and IFS ENS. The rank histograms of ArchesWeatherGen are much flatter than the other methods at 1 day lead time. At 7 days lead time, its rank histograms are better than NeuralGCM for Z500 and Q700 but slightly worse for the other variables due to remaining under-dispersiveness.}
    \label{fig:rank_histograms}
\end{figure}

To validate the distributions of our generated ensembles, we use rank histograms \cite{talagrand1999evaluation}, which consists of computing the rank of the ground truth among the $M$-members ensemble for each sample, and then calculating the histograms of these ranks, averaging over spatial locations. For perfect ensemble forecasts, we expect flat rank histograms. Overdispersive models are characterized by $\cap$-shaped rank histogram, and under-dispersive models by a $\cup$-shaped rank histogram. In Figure \ref{fig:rank_histograms}, we compare the rank histograms of ArchesWeatherGen, NeuralGCM and IFS ENS on upper-air variables. At 24h, ArchesWeather almost perfectly flat histograms (very slightly underdispersive for the extremes), much better than IFS ENS and NeuralGCM. At 7 days lead time, the rank histograms are not flat anymore, and slightly under-dispersive. Overall, they are flatter than IFS ENS, better than NeuralGCM for Z500 an Q700 and slightly worse on T850, U850 and V850 due to the remaining underdispersiveness. We believe that this could be solved by adjusting the noise scaling parameter separately for each physical variable, which we did not investigate for simplicity of the approach.

\subsection{Ablations of the generative model} \label{section:gen_ablation}

In this section, we validate our design choices for ArchesWeatherGen.

\paragraph{Impact of OOD fine-tuning and noise scaling. } As explained in Section \ref{overfitting_underdispersion}, overfitting of the underlying deterministic model results in test residuals that are slightly "out-of-distribution" compared to residuals on the train set, and in particular have a higher norm than on the train set, which results in under-dispersion. We now evaluate the effectiveness of our strategies to mitigate this problem: fine-tuning the generative model on year 2019 (on which the deterministic model has not been trained) and scaling the variance of the noise given as input to the generative model.

\begin{figure}[ht]
     \centering
     \includegraphics[width=\linewidth]{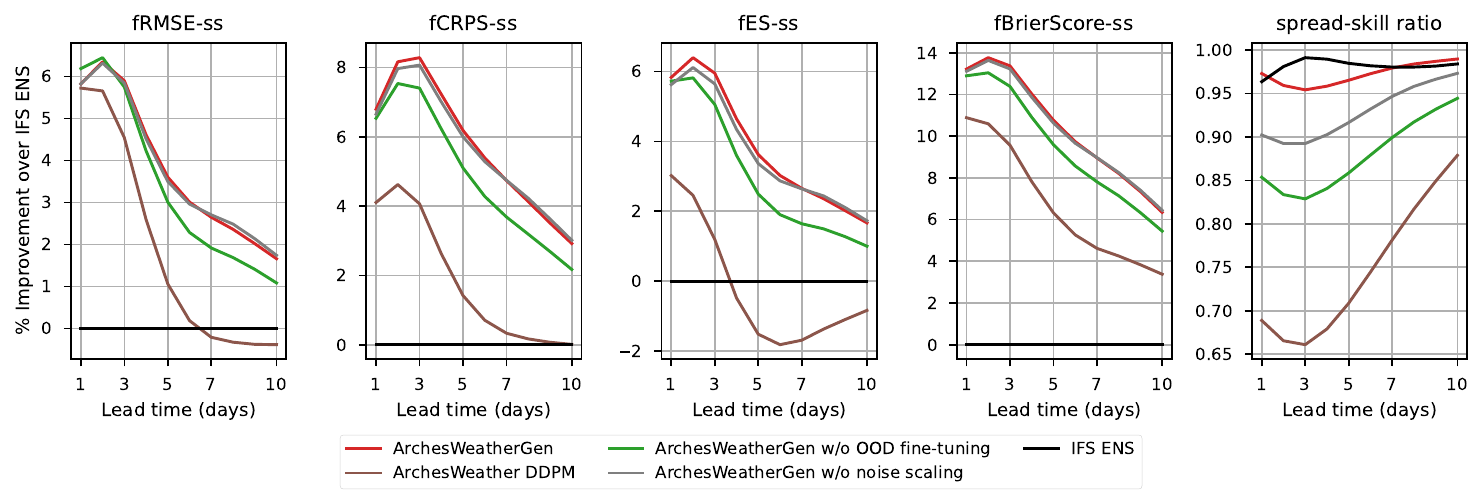}
     \caption{Ablation of our strategies to improve ensemble metrics and dispersion of residual generative models. OOD fine-tuning helps to improve all ensemble metrics, while noise scaling only reduces under-dispersion.}
     \label{fig:main_ablation}
 \end{figure}

In Figure \ref{fig:main_ablation}, we first see that our flow matching model greatly improves upon the DDPM variant for all ensemble metrics but notably (and perhaps unexpectedly) for spread-skill ratio: at 24h, the DDPM version has a spread-skill ratio of 0.68, compared to 0.85 for the basic flow matching version without OOD fine-tuning or noise scaling. 
Building upon this base version, we add OOD fine-tuning which improves all metrics and has a better 24h spread-skill ratio of 0.9. Finally, with noise scaling on top, the ensemble metrics are not significantly different, but the spread-skill ratio is improved to around 0.96 and increases to 0.98 at longer lead times. Overall, this experiment shows that fine-tuning helps combat distribution mismatch and improve all ensemble metrics. On the other hand, noise scaling has no effect on RMSE, CRPS or BrierScore, but helps to improve the dispersiveness of our model. 

We also tried a more naive way of increasing the dispersion of the model, which is to rescale the output of the residual generative model. Although this was effective in adjusting the spread-skill ratio, the other ensemble metrics (and notably CRPS) were degraded, which is expected since the probability distribution of residuals with a higher norm is a priori different from the unconditional distribution of residuals. On the other hand, a higher-norm input noise is guided by our flow matching model towards the true probability distribution of higher-norm residuals, as evidenced by the higher dispersion obtained with similar or slightly higher CRPS scores.

\paragraph{Impact of deterministic model.}

\begin{figure}[t]
    \centering
    \includegraphics[width=\linewidth]{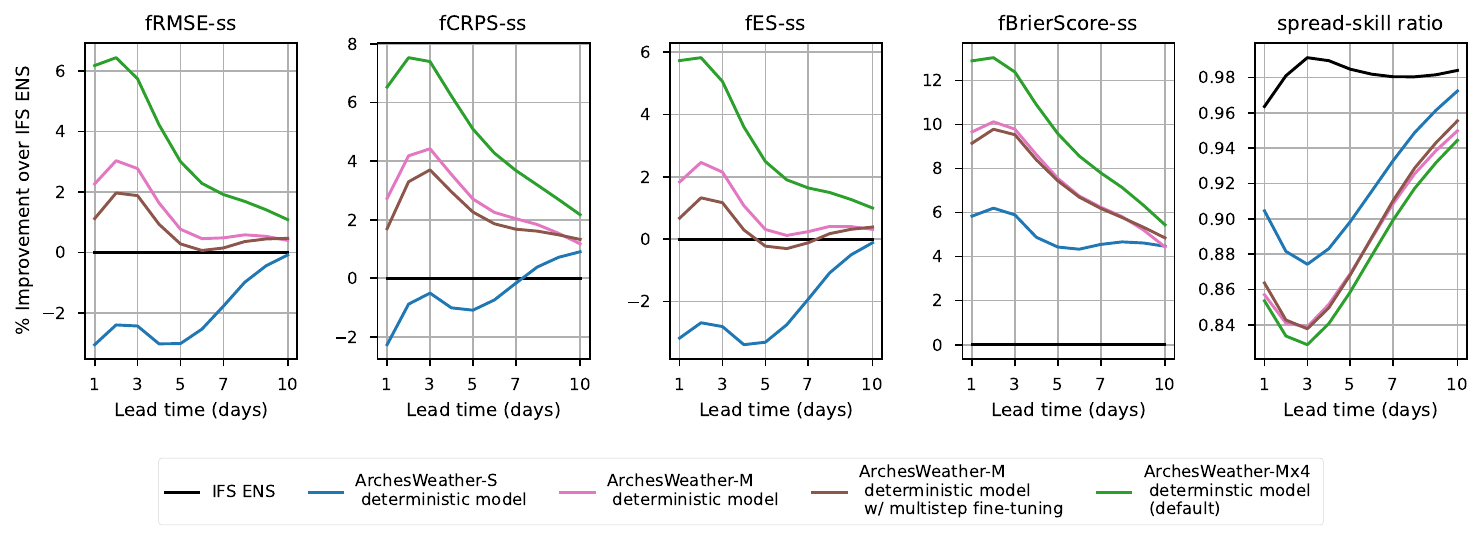}
    \caption{Impact of the choice of deterministic model used in residual modelling on ensemble metrics. Better deterministic models improve all metrics, especially at shorter lead times.}
    \label{fig:det_model_ablation}
\end{figure}

Our residual generative models require a choice of deterministic model to compute the residuals on which it is trained.
In Figure \ref{fig:det_model_ablation}, we evaluate the impact of this design choice on the performance of our flow matching models.
We use four different deterministic models: ArchesWeather-S, ArchesWeather-M, ArchesWeather-M with multi-step fine-tuning, and ArchesWeather-Mx4, the latter being the one used in ArchesWeatherGen.

We can see that the choice of underlying deterministic model has a big impact on ensemble metrics, with bigger deterministic models yielding better residual generative models. This means that using better deterministic models than ArchesWeather-Mx4 could potentially bring further gains. One small downside of using larger deterministic models is that the spread-skill ratio is slightly decreased, which is another manifestation of overfitting, but we have proposed solutions to handle this problem.
Finally, an interesting observation is to compare a deterministic model trained with next-step prediction only versus one that has been fine-tuned on auto-regressive rollouts. Using the fine-tuned model seems to decrease performance of our generative model, probably because it is slightly worse for next-step prediction compared to the model without multi-step fine-tuning.

\subsection{Qualitative Analysis on Hurricane Teddy}

\begin{figure}[ht]
    \centering
    \includegraphics[width=\linewidth]{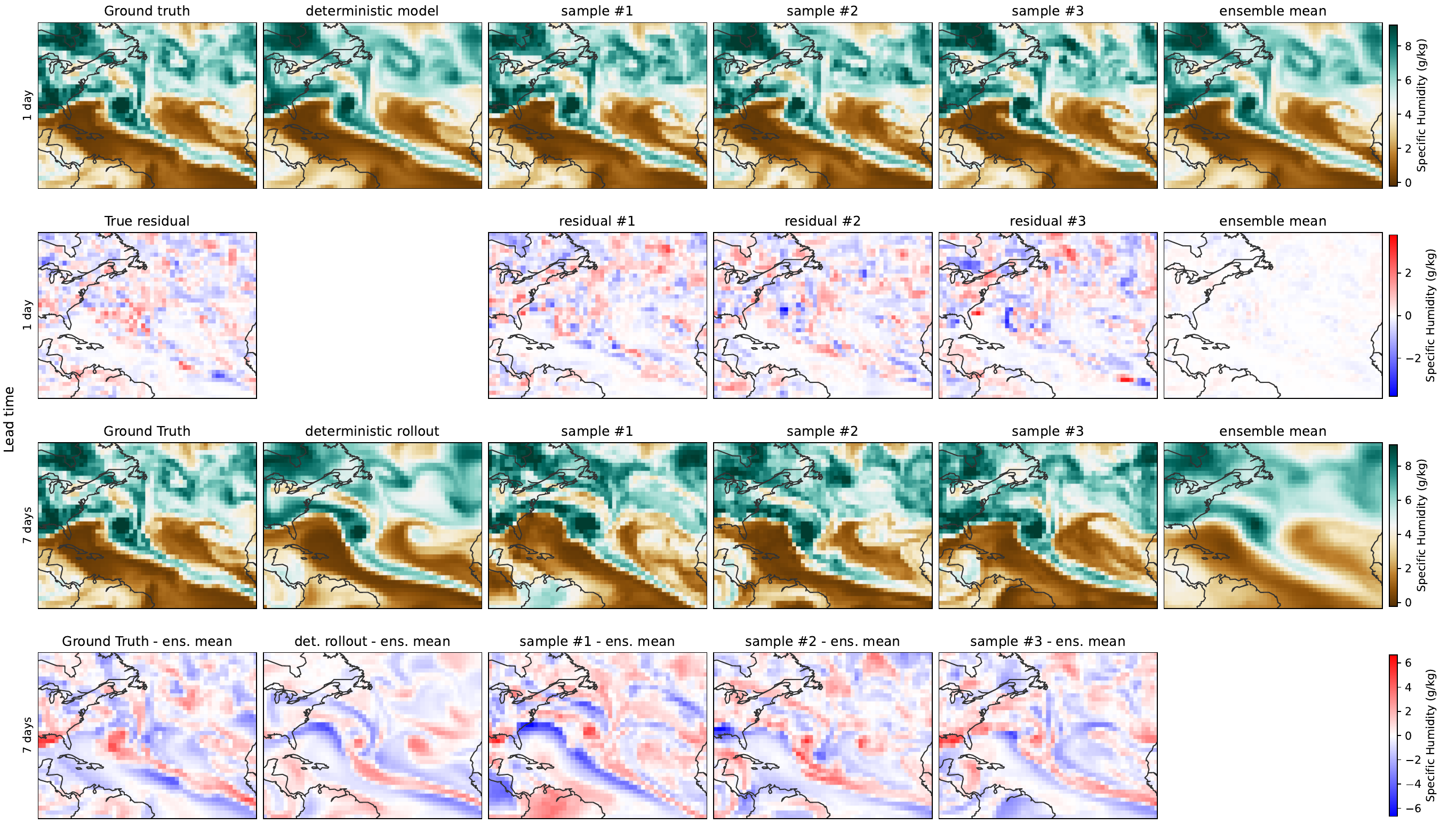}
    \caption{Visualization of 700hPa specific humidity predictions made by our generative weather model on Hurricane Teddy at 12pm on the 21th of September, 2020. Predictions are made from one day before this date (two first rows) and seven days before this date (two last rows). On the second row, we visualize the residuals generated by our flow matching model, which are added to the one-day deterministic prediction to recover the one-day ensemble predictions. One the last row, we visualize the difference between samples and the ensemble mean, computed from a 50-members ensemble. The deterministic predictions are made by our deterministic model, and are blurrier than the ground truth.}
    \label{fig:teddy}
\end{figure}

In Figure \ref{fig:teddy}, we show humidity samples generated by our model for Hurricane Teddy on September 21, 2020.

Hurricane Teddy was the fifth-largest Atlantic hurricane by diameter of gale-force winds recorded. It produced large swells along the coast of the eastern United States and Atlantic Canada. It resulted in massive damages over the island of Bermuda on the 21st of September when the storm suddenly intensified. In the first line of the figure, we present the cyclone as reanalyzed in ERA5, and contrast it with the 1-day ahead forecast of the deterministic model (ArchesWeather), some samples of the 50-member generated ensemble of ArchesWeatherGen, and their ensemble mean. As expected, the ensemble mean is extremely close to the deterministic prediction. However, we can observe some small-scale variations on the individual samples, such as the convection plume splitting in sample two, and slight displacements of the small-scale structures, corresponding to slight variations in the trajectory of the hurricane. These can be observed in the shape of the errors, generally corresponding to small differences in the advection of water masses. In all these cases, the forecast passes very close to Bermuda.

If we examine the 7-day rollouts of both the deterministic model and the generative model, all models predict the intensification from a much weaker version of the cyclone on September 14 (their initial condition). We can notice a much higher variance in the generated samples, corresponding to different possible trajectories. This is reflected in the smoothness of the ensemble mean, corresponding to averaging different larger-scale structures.

Individual trajectories obtained with the generative approach contain sharp features compared to the deterministic rollout, which becomes smooth at a 7-day lead time. This sample sharpness is further illustrated by the clear large-scale structures that emerge when visualizing the differences between the individual members and the ensemble mean, seen on the fourth line of the figure. This corresponds to different trajectories that the hurricane could have followed and variations in its intensity. We can also observe the formation or dissipation of a secondary, weaker, cyclonic structure eastwards of Hurricane Teddy in the 7-day rollouts. Multiple of the generated trajectories remain close to the actual path of the hurricane. Twenty-five more samples can be visualized in the Appendix \ref{fig:teddy_samples}.

\subsection{Discussion and future work}

\paragraph{Learning spherical data with transformers.} Our data source is the classical equirectangular projection of the ERA5 dataset, which presents some modeling challenges: physical phenomena are distorted near the poles, requiring faster information exchange in the neural network. We tried accounting for the sphere topology in two ways: first, we experimented with a shifting mechanism in Swin Transformers that connects some parts of the northern data that are otherwise very distant in the latitude-longitude representation. This did not improve the forecasting error since at 1.5º, information exchange in attention windows covers long distances, and any two regions on Earth can exchange information throughout the model. However, in ArchesWeather, as in Pangu-Weather, we do connect the left-most (-180ºW) and right-most (180ºE) image parts at every layer, making our data and network effectively cylindrical. 

Second, we tested weather forecasting on a cubed Earth projection, with a stack of 6 faces and attention windows that can exchange information across neighboring faces. While this seems promising, we did not pursue further due to two hurdles: (i) The 1.5º data is downsampled from the 0.25º equirectangular projection, which results in the 1.5º data having some high frequencies in the high latitudes than cannot be obtained from projecting a 1.5º cubed Earth map. (ii) In WeatherBench, the reference data for computing the forecast error is the equirectangular map at 1.5º. Training a model on another representation, like cubed Earth, exposes the model to an additional reprojection error when going back to the latitude-longitude format for computing metrics in WeatherBench. Therefore, the cubed Earth neural network needs to be trained with an MSE loss on the equirectangular map, requiring a differentiable projection after the transformer layers to convert the cubed Earth format. This differentiable projection would allow the neural network to compensate for the reprojection error. These two problems are not major blockers and we believe that at 0.25º and a differentiable projection back to the equirectangular map, using transformers on a cubed Earth representation could potentially bring improvements, as it did for Convolutional Neural Networks \cite{weyn2020improving}.

\paragraph{Overfitting and under-dispersion.} We have presented a methodology for training diffusion-based generative models, based on removing the deterministic component with a deterministic model trained with the root mean square error. Overfitting of the base deterministic model gives smaller residual data, on which the generative model is trained. With a similar spread between train and test for the generative models and worse Ensemble Mean RMSE on the test set compared to the train set, the test spread-skill ratio is lower than on the train set, resulting in under-dispersion. To reduce this effect, we fine-tune our model on data from 2019, on which the deterministic model is not trained. However, more gains could potentially be obtained by allocating more training data for this fine-tuning stage, which means training the deterministic model on fewer data. However, any such gains from fine-tuning the generative on more data could be more than compensated for by a deterministic model less skilled, since trained on less data. Determining the optimal data split is left for future work.

\paragraph{Non-Gaussian residual data.} By removing the expectation component learned by deterministic models, the data might not follow a Gaussian distribution anymore, which we have confirmed experimentally: the tails are much heavier than for normally distributed data, with frequent outliers that have a high value after unit variance rescaling. This might make diffusion modeling harder \cite{pandey2024heavy}. We have tried to map this data to a Gaussian distribution via a contraction mapping $f$ (e.g. square root or log), modeling $f(x)$ and then applying $f^{-1}$ to map back to a real sample. Although this improved metrics for the first few autoregressive steps, mapping back with $f^{-1}$ gave us instabilities in the rollouts. We did not pursue this further and leave this for future work.

\paragraph{Multi-step fine-tuning.} We have approached the problem of generating consistent trajectories by modeling the transition function and then composing it auto-regressively, which relies on a Markovian assumption of the underlying data/physical phenomenon. While this assumption seems to hold well in practice, its limitations could theoretically be addressed by fine-tuning the model on auto-regressive rollouts. Fine-tuning could also help to correct small modeling defects in the one-step transition function and improve multi-step predictions. Although this strategy is beneficial for deterministic models, we found in Figure \ref{fig:det_model_ablation} that the use of multi-step fine-tuned models as base models in our framework is harmful, because it slightly degrades the quality of the transition model. However, the gap between the two methods reduces to 0 as lead time increases, meaning that either (i) the bias introduced by multi-step fine-tuning is less and less important over time, or (ii) multi-step fine-tuning slightly helps for generative rollouts but it is more than compensated by the worse model metrics at 24h, which could be addressed with a better multi-step fine-tuning procedure. We leave investigating this for future work.

Another possibility is to fine-tune the generative model itself from its rollouts. However, this requires us to fine-tune models through many diffusion sampling steps, whose naive implementation is impracticable. 

\paragraph{Diffusion modeling and initial condition perturbation.} In this work, we do not consider perturbations of the initial condition (IC) to generate weather trajectories. In theory, even if there is some uncertainty in the true weather state $\x_t$ that affects the weather trajectories, the learned transition distribution $p(\x_{t+\delta}|\x_t)$ should take this uncertainty into account, unlike traditional physical models.
In practice, the GenCast model uses diffusion modeling, as well as initial condition perturbation derived from the ERA5 Ensemble of Data Assimilation (EDA) \cite{isaksen2010ensemble}. The authors show that, without perturbation, the diffusion model is underdispersive. Although we have also observed underdispersiveness for our model, we have proposed other solutions that are simpler to implement compared to IC perturbation and solve the underdispersion issue. This might indicate that initial condition perturbation is not needed to generate weather trajectories with the correct dispersion. However, using an ensemble of IC conditions does provide more information about the uncertainties of weather conditions than a single state and could potentially improve the skill of the model.

\section{Conclusion}

We have presented ArchesWeather and ArchesWeatherGen, two machine learning models for weather prediction that operate at 1.5º resolution. ArchesWeather is a deterministic model trained with a 24h lead time, and has very good skill thanks to various neural network architecture innovations. ArchesWeatherGen is a flow matching model with the same backbone architecture, trained to project predictions from ArchesWeather to the full distribution of ERA5 weather states, enabling probabilistic forecasting. We find that ArchesWeatherGen surpasses IFS ENS and even NeuralGCM on most physical variables, as measured by probabilistic ensemble metrics, including Ensemble Mean RMSE, CRPS and Brier score. We have found that a main obstacle to unlocking the full potential of residual modeling is overfitting of the underlying deterministic model. Fine-tuning the model on data that the deterministic model has not been trained on and scaling the initial noise of the sampling process were effective strategies to tackle this problem. Our best model can be trained with a computational budget of $~23$ A100 days, paving the way for more accessible research on ML weather forecasting.

Our generative model not only provides accurate future weather trajectories but is also naturally suited for data assimilation. In fact, diffusion-based models (including flow matching models) can be used to predict weather states for which we have measurements and want to sample the conditional distribution with respect to these measurements, for example, with score-based data assimilation \cite{rozet2023score, manshausen2024generative}. Operating at 1.5º, our models are, however, less suited for applications that require a better resolution, like cyclone tracking, or regional forecasting. The output of our models could potentially be downscaled to a finer resolution, which we leave for future work.

\section{Acknowledgements}
We thank Robert Brunstein for evaluating the 2D transformer version of ArchesWeather. Many thanks to Clément Deauvilliers for experimenting with cyclone tracking at 1.5º with ArchesWeather. Thanks to Graham Clyne and David Landry from the ARCHES team for fruitful discussions on diffusion modeling. G.C., R.S., and C.M. were supported by the French government, via the Choose France Chair in AI. This work was granted access to the HPC resources of IDRIS under the allocation AD010114684 made by GENCI.

\newpage


%% file: figs/weatherbench_deterministic.tex
\begin{table}[hb]
  \centering
  \setlength{\tabcolsep}{4pt}
\begin{small}
\begin{sc}
\begin{tabular}{lllccccccccc}
\toprule
                  & \textbf{Res.} & \textbf{Cost} & \begin{tabular}{@{}c@{}}\textbf{Z500} \\ $m^2/s^2$ \end{tabular} & \begin{tabular}{@{}c@{}}\textbf{T850} \\ $ºK$ \end{tabular} & \begin{tabular}{@{}c@{}}\textbf{Q700} \\ $g/kg$ \end{tabular} & \begin{tabular}{@{}c@{}}\textbf{U850} \\ $m/s$ \end{tabular} & \begin{tabular}{@{}c@{}}\textbf{V850} \\ $m/s$ \end{tabular} & \begin{tabular}{@{}c@{}}\textbf{T2m} \\ $ºK$ \end{tabular} & \begin{tabular}{@{}c@{}}\textbf{SP} \\ $Pa$ \end{tabular} & \begin{tabular}{@{}c@{}}\textbf{U10m} \\ $m/s$ \end{tabular} & \begin{tabular}{@{}c@{}}\textbf{V10m} \\ $m/s$ \end{tabular} \\
\midrule
IFS HRES               &               &                                                                & 42.30         & 0.625         & 0.556         & 1.186         & 1.206         & 0.513        & 60.16       & 0.833        & 0.872        \\
\midrule
Pangu-Weather             & 0.25º         & 2880                                                           & 44.31         & 0.620         & 0.538         & 1.166         & 1.191         & 0.570        & 55.14       & 0.728        & 0.759        \\
NeuralGCM         & 0.25º         & 16128                                                          & 37.94         & 0.547         & 0.488         & 1.050         & 1.071         & N/A          & N/A         & N/A          & N/A          \\
FuXi              & 0.25º         & 52                                                             & 40.08         & 0.548         & N/A           & 1.034         & 1.055         & 0.532        & 49.23       & 0.660        & 0.688        \\
GraphCast         & 0.25º         & 2688                                                           & 39.78         & 0.519         & 0.474         & 1.000         & 1.02          & 0.511        & 48.72       & 0.655        & 0.683        \\
\midrule
Keisler           & 1º            & 11                                                             & 66.87         & 0.816         & 0.658         & 1.584         & 1.626         & N/A          & N/A         & N/A          & N/A          \\
SphericalCNN      & 1.4º          & 384                                                            & 54.43         & 0.738         & 0.591         & 1.439         & 1.471         & N/A          & N/A         & N/A          & N/A          \\
Stormer           & 1.4º          & 256                                                            & 45.12         & \bfu{0.607}         & 0.527         & \bfu{1.138}         & \bfu{1.156}         & 0.570        & \textbf{53.77}       & \bfu{0.726}        & \bfu{0.760}        \\
NeuralGCM ENS (50) & 1.4º          & 7680                                                           & \bfu{43.99}         & 0.658         & 0.540         & 1.239         & 1.256         & N/A          & N/A         & N/A          & N/A          \\
\midrule
\midrule
ArchesWeather-S & 1.5° & 9 & 48.92& 0.650& 0.542& 1.289& 1.325 & 0.562& 60.30& 0.833& 0.873\\
ArchesWeather-M & 1.5° & 18 & 44.96& 0.615& \textbf{0.527} & 1.219& 1.251& \textbf{0.539}& 56.22& 0.783& 0.819\\
ArchesWeather-Mx4 & 1.5° & 36 & \bfu{41.93} & \bfu{0.593} & \bfu{0.513} & \textbf{1.172} & \textbf{1.203} & \bfu{0.517} & \bfu{52.22} & \textbf{0.749} & \textbf{0.783} \\
\bottomrule

\end{tabular}
\end{sc}
  \vspace{1em}

\end{small}
  \caption{Comparison of deterministic ML weather models on RMSE scores for key weather variables at a 24h lead time. Cost is the training computational budget in V100-days. Best scores for training resolution coarser than 1º are in \bfu{underlined bold}, second best scores in \textbf{bold}.}
  \label{full_results}
\end{table}

%% file: figs/ablation_table.tex
\begin{table}[ht]
  \centering
\begin{small}

\begin{tabular}{l|ccccc|ccc}
\toprule
  \sc{\textbf{Model}} & \begin{tabular}{@{}c@{}}Cross-Level \\ Attention \end{tabular} & \begin{tabular}{@{}c@{}}Recent Past \\ Finetuning \end{tabular} & SwiGLU & Ensemble & \# Layers & \sc{\textbf{Z500↓}} & \sc{\textbf{T2m↓}} & \sc{\textbf{RMSE-ss↑}} \\
 
\midrule
Pangu-S & \xmark & \xmark & \xmark & \xmark  & 16 & 66.7 & 0.84 & -30.6 \\
Base improvements & \xmark & \xmark & \xmark & \xmark & 16 & 55.1 & 0.594 & -17.1 \\
+ Cross-Level Attention & \cmark & \xmark & \xmark & \xmark & 16 & 50.6 & 0.567 & -9.7 \\
+ 2007-2018 fine-tuning & \cmark & \cmark & \xmark & \xmark & 16 & 49.3 & 0.566 & -8.6 \\
ArchesWeather-S & \cmark & \cmark & \cmark & \xmark & 16 & 48.92 & 0.562 & -5.2 \\

\midrule
Pangu-M & \xmark & \xmark & \xmark & \xmark  & 32 & 58.7 & 0.78 & -20.4 \\
Base improvements & \xmark & \xmark & \xmark & \xmark & 32 & 51.8 & 0.572 & -11.6 \\
+ Cross-Level Attention & \cmark & \xmark & \xmark & \xmark & 32 & 48.7 & 0.552 & -5.6  \\
+ 2007-2018 fine-tuning & \cmark & \cmark & \xmark & \xmark & 32 & 48.0 & 0.551 & -5.0  \\
ArchesWeather & \cmark & \cmark & \cmark & \xmark & 32 & 44.96 & 0.539 & 0.66 \\
ArchesWeather-Mx4 & \cmark & \cmark & \cmark & \cmark & 32 & 41.93 & 0.517 & 5.3 \\

\midrule
2D ArchesWeather & \xmark & \cmark & \cmark & \xmark  & 32 & 53.0 & 0.575 & -8.0 \\

\bottomrule
\end{tabular}
\vspace{1em}
\end{small}
\caption{500hPa geopotential and 2m temperature RMSE at 24h lead-time for different version of our models, and Pangu-Weather re-trained at 1.5º. RMSE-ss is the relative RMSE improvement over HRES. Models without our proposed Cross-Level Attention module (CLA) use local 3D attention, except the 2D ArchesWeather that uses traditional 2D attention with vertical information stacked along the embedding dimension.}
\label{abl}
\end{table}

%% file: sections/appendix.tex
\appendix

\section{Appendix: Additional details}
\subsection{Training details}\label{app:training-details}

We denote $(\x_t)_{t \in \mathcal{D}}$ the historical trajectory of ERA5, indexed by time $t$. Input states $\x_t$ are normalized to zero mean and unit variance on a per-variable and per-level basis, using statistics of the training set 1979-2018. 

Following GraphCast, we scale the training loss with coefficients proportional to the air density, to give more importance to variables closer to the surface. We also use the same reweighting of the surface variables with a coefficient of $1$ for 2m temperature, and $0.1$ for wind components and mean surface pressure.

We train all our models with the AdamW optimizer \cite{kingma2014adam}. The batch size is $4$ and the optimizer parameters are a learning rate of 3e-4, beta parameters $(\beta_1=0.9, \beta_2=0.98)$ and a weight decay of $0.05$. The learning rate is increased linearly for the first 5000 steps, then decayed with a cosine schedule for the remaining steps.

\subsection{Comparison with state-of-the-art} \label{app:sota}

For deterministic models (except Stormer), RMSE scores at 1.5º are taken from WeatherBench2 \cite{rasp2023weatherbench}. To compute the training costs associated to each method, we use the following ratios to compare GPU cards: 1 A100 = 2 V100, 1 H100 = 4 V100, 1TPUv4 = 3 V100, 1 TPUv5e = 6 V100.

For Stormer \cite{nguyen2023scaling}, we evaluate outputs provided by the authors at 1.4º resolution. Stormer is a $\sim$300M parameters model trained to forecast ERA5 variables at multiple lead times simultaneously: 6h, 12h and 24h. To make a 24h leadvtime forecast, Stormer uses all possible combinations of lead times as conditioning: 24h, 12h-12h, 12h-6h-6h, 6h-12h-6h, 6h-6h-12h, 6h-6h-6h-6h, and averages all trajectories. This base model is run 16 times with different lead time conditioning to make a 24h forecast. Our ensemble model requires four model forwards with $\sim336M$ total parameters. Please see the paper \cite{nguyen2023scaling} for more details on Stormer.

\section{Additional experiments}

\subsection{Impact of OOD fine-tuning and noise scaling}

\begin{figure}[ht]
    \centering
    \includegraphics[width=\linewidth]{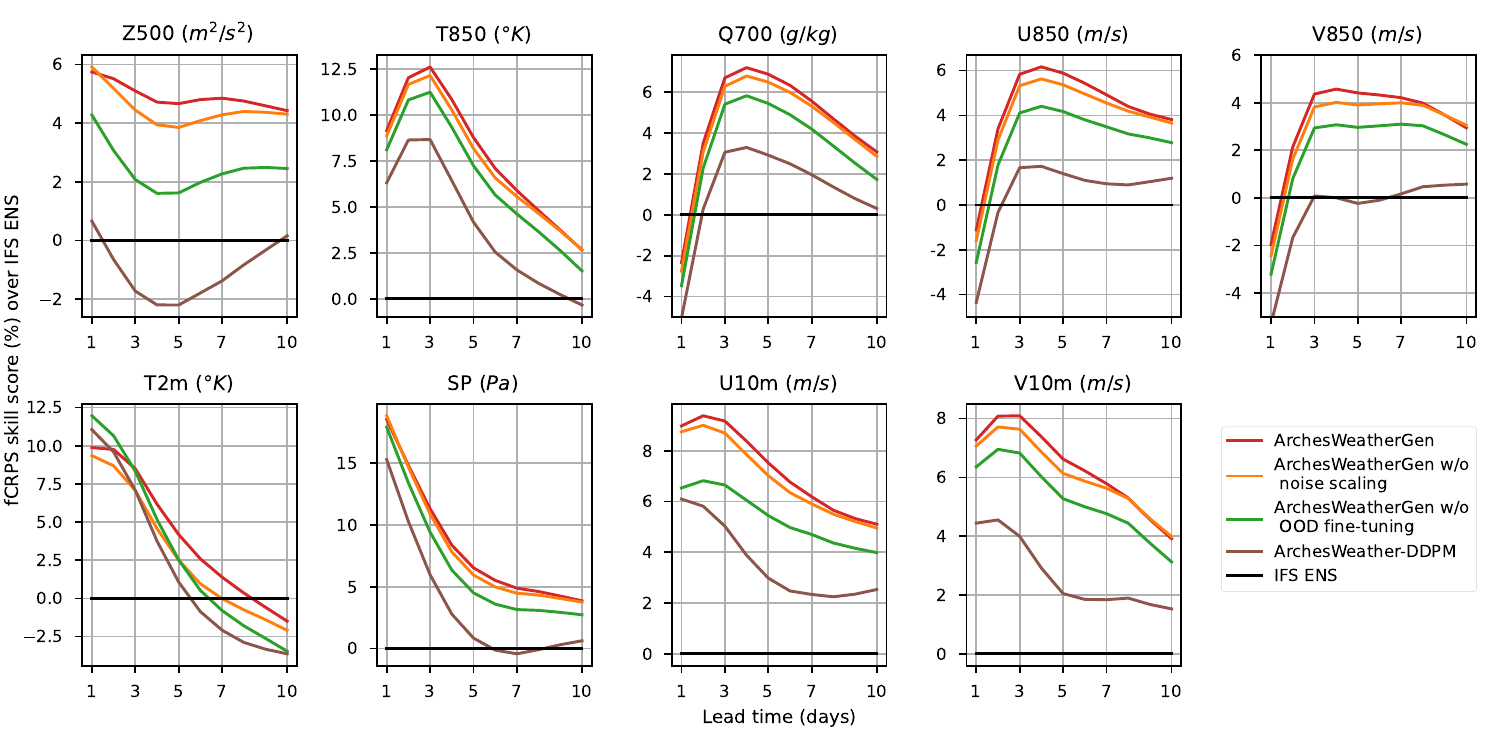}
    \caption{fCRPS scores per headline variable.}
    \label{fig:variable_rmse}
\end{figure}

In figure \ref{fig:variable_rmse}, we report the CRPS scores of various methods separately for each headline variable in WeatherBench. We can see that the ranking of methods is the same for all variables, with ArchesWeatherGen having the best scores and ArchesWeather-DDPM the lowest.

\begin{figure}[ht]
    \centering
    \includegraphics[width=\linewidth]{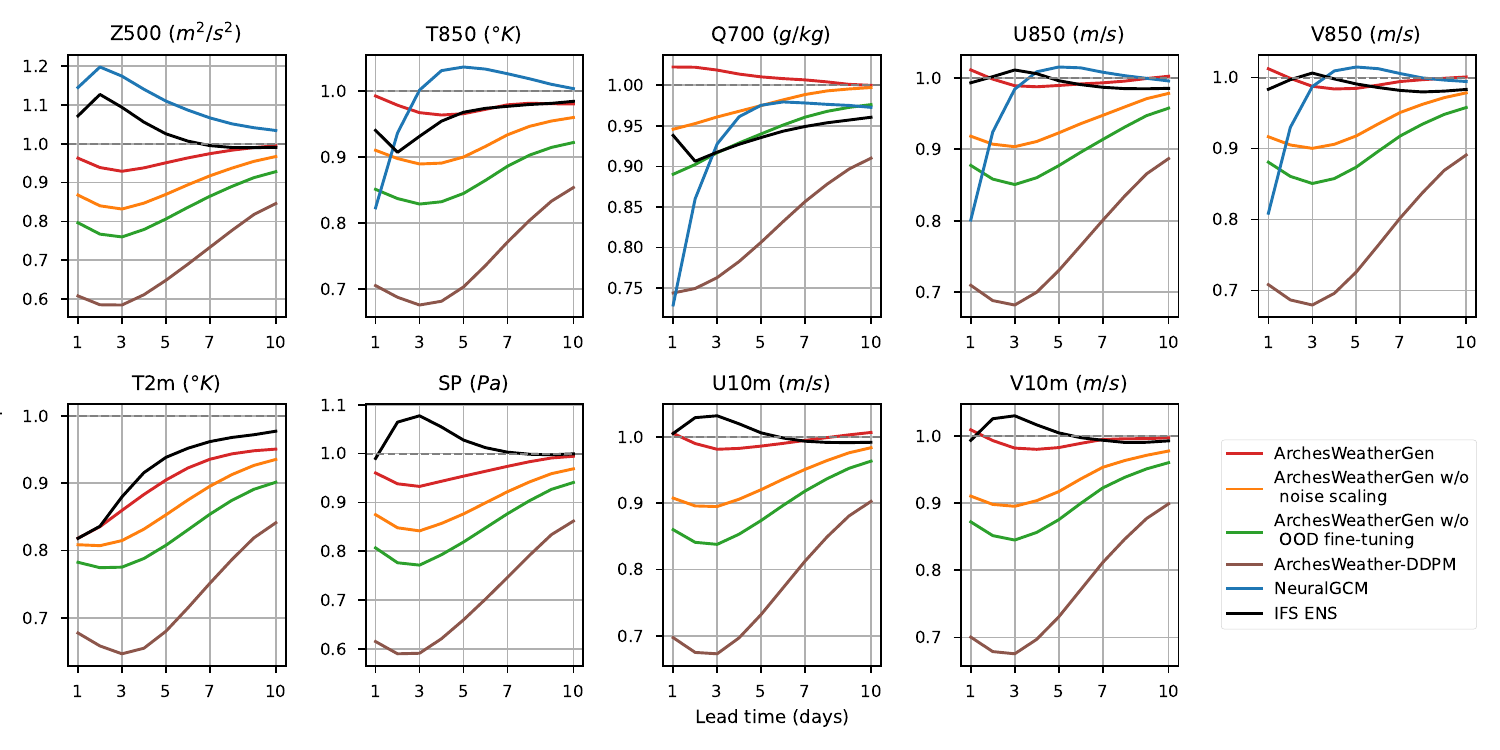}
    \caption{Spreak-skill ratios of our methods with different variables. Using OOD and noise scaling helps to recover correct dispersiveness across lead times.}
    \label{fig:variable_spskr}
\end{figure}

In figure \ref{fig:variable_spskr}, we report the spread-skill ratio of our methods separately for each headline variable in WeatherBench. Again, we can see that the qualitative behavior is the same for all variables. Using noise scaling helps to reach a spread-skill ratio close to 1 for all lead times, except for 2m-temperature.

\section{Additional visualizations}

\subsection{Samples generated on Hurricane Teddy}

\begin{figure}[ht]
    \centering
    \includegraphics[width=\linewidth]{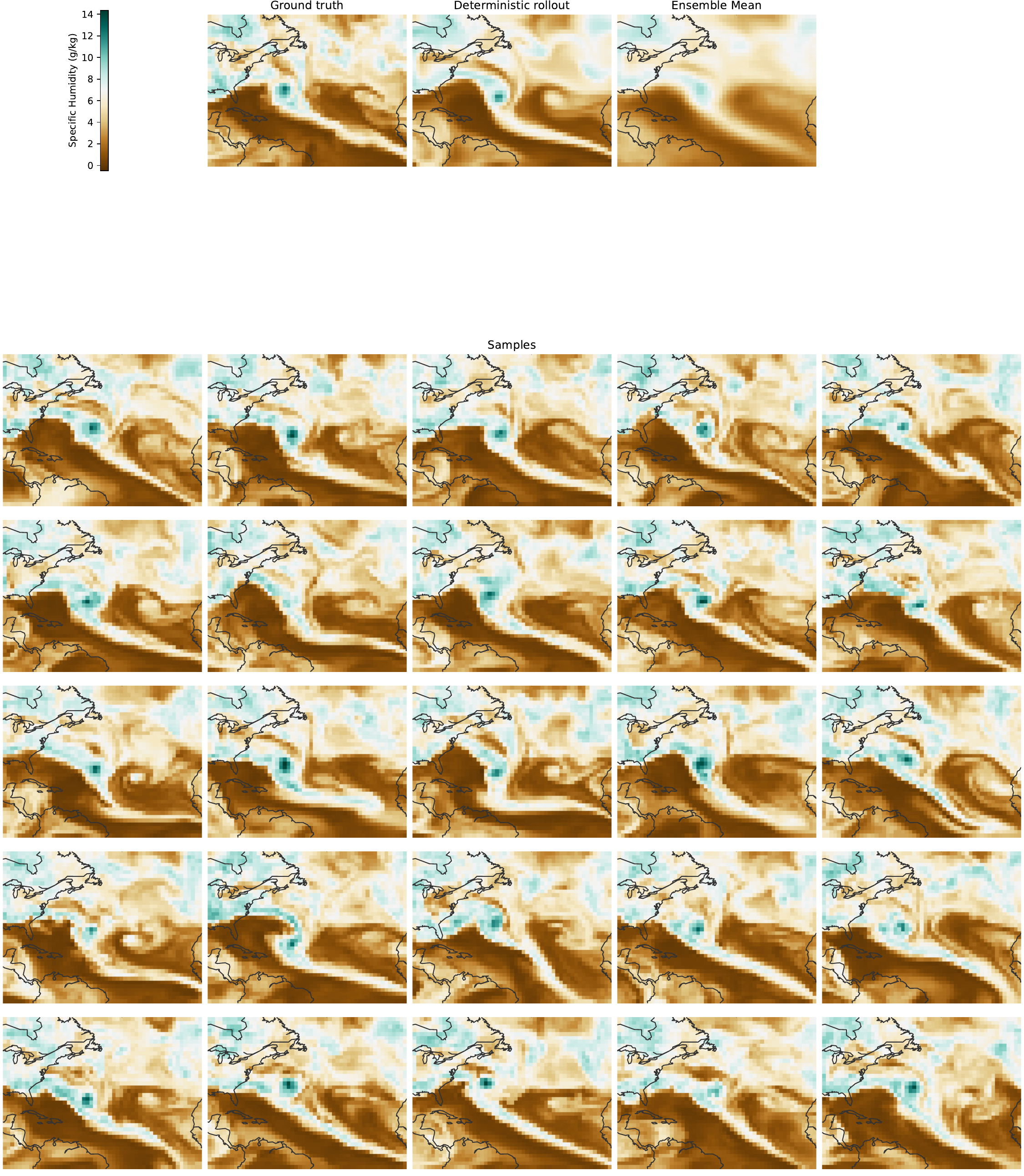}
    \caption{25-member 7-day forecast generated by ArchesWeatherGen on Hurricane Teddy, initialized September 14th, 2020.}
    \label{fig:teddy_samples}
\end{figure}

In figure \ref{fig:teddy_samples} we can see 25 generations of a 7-days rollout of ArchesWeatherGen. Of these, we can notice that the hurricane structure dissipated in three of them, with the rest having different degrees of intensification. The trajectories are physically realizable, and most of them follow trajectories that would lead to accurate warnings of an intensification of the hurricane, giving ample warning time.